\documentclass[acmlarge]{acmart}
\usepackage{amsmath,amsfonts,amsthm}
\usepackage{amssymb}
\usepackage{mathtools,nccmath}
\usepackage{mathrsfs}
\usepackage[compact]{titlesec}

\usepackage{subfigure}
\usepackage{caption}
\usepackage{booktabs}
\usepackage{multirow}
\usepackage{url}
\usepackage{numprint}
\usepackage{ragged2e}
\usepackage{tabularx}
\usepackage{natbib} 
\usepackage[inline]{enumitem}
\usepackage{tikz}
\usepackage{xcolor}

\newcommand{\METHOD}{\texttt{COCOA}}
\newcommand{\Barlow}{Barlow Twins}

\AtBeginDocument{%
  \providecommand\BibTeX{{%
    \normalfont B\kern-0.5em{\scshape i\kern-0.25em b}\kern-0.8em\TeX}}}

\setcopyright{acmcopyright}
\acmJournal{IMWUT}
\acmYear{2022} \acmVolume{6} \acmNumber{3} \acmArticle{108} \acmMonth{9} \acmPrice{15.00}\acmDOI{10.1145/3550316}

\begin{document}

\title{\METHOD: Cross Modality Contrastive Learning for Sensor Data}

\author{Shohreh Deldari}
\email{shohreh.deldari@student.rmit.edu.au}
\orcid{0000-0001-8150-120X}
\affiliation{%
  \institution{School of Computing and Technologies, RMIT University}
  \city{Melbourne}
  \state{Victoria}
  \country{Australia}
  \postcode{3000}
}

\author{Hao Xue}
\affiliation{%
  \institution{School of Computer Science and Engineering, University of New South Wales}
  \streetaddress{}
  \city{Sydeny}
  \state{NSW}
  \country{Australia}}
\email{hao.xue1@unsw.edu.au}
\orcid{0000-0003-1700-9215}

\author{Aaqib Saeed}
\affiliation{%
  \institution{Philips Research}
  \city{Eindhoven}
  \country{Netherlands}
}
\orcid{0000-0003-1473-0322}
\email{aaqib.saeed@philips.com}
\author{Daniel V. Smith}
\affiliation{%
  \institution{Data61, CSIRO}
  \city{Hobart}
  \state{Tasmania}
  \country{Australia}}
\email{daniel.v.smith@csiro.au}
\orcid{0000-0001-9983-095X}

\author{Flora D. Salim}
\affiliation{%
  \institution{School of Computer Science and Engineering, University of New South Wales}
  \city{Sydney}
  \state{NSW}
  \country{Australia}
  \postcode{2000}}
\email{flora.salim@unsw.edu.au}
\orcid{0000-0002-1237-1664}

\renewcommand{\shortauthors}{Deldari, et al.}

\begin{abstract}
 Self-Supervised Learning (SSL) is a new paradigm for learning discriminative representations without labeled data, and has reached comparable or even state-of-the-art results in comparison to supervised counterparts. Contrastive Learning (CL) is one of the most well-known approaches in SSL that attempts to learn general, informative representations of data. CL methods have been mostly developed for applications in computer vision and natural language processing where only a single sensor modality is used. A majority of pervasive computing applications, however, exploit data from a range of different sensor modalities.  While existing CL methods are limited to learning from one or two data sources, we propose \METHOD\ (Cross mOdality COntrastive leArning), a self-supervised model that employs a novel objective function to learn quality representations from multisensor data by computing the cross-correlation between different data modalities and minimizing the similarity between irrelevant instances. 
We evaluate the effectiveness of \METHOD\ against eight recently introduced state-of-the-art self-supervised models, and two supervised baselines across five public datasets. We show that \METHOD\ achieves superior classification performance to all other approaches. Also, \METHOD\ is far more label-efficient than the other baselines including the fully supervised model using only one-tenth of available labeled data.
\end{abstract}



\maketitle

\section{Introduction}
\label{sec:intro}

\subsection{Why Self-Supervised Learning?}
Supervised deep learning models have achieved great success in recent decades; however, their success heavily depends on the accuracy and existence of annotated data. Collecting labelled datasets and annotating a huge amount of data from a variety of input sources is not a feasible task or at least is a time-consuming and expensive process which results in high-quality labelled datasets being very limited. Some domains, such as medicine, law, biology, etc, require domain experts for annotating the collected data. In some cases, annotation can raise privacy issues specially in medical and other human-related applications.  

Recently, Self-Supervised Learning (SSL) techniques have emerged as a valuable alternative to the supervised learning paradigm to exploit large-scale unlabeled data without requiring manual annotations. Self-supervised learning methods generates and consumes self-defined labels to capture important features of data. The main advantage of pre-training models in a self-supervised manner is that the SSL models do not require annotated data. Hence they can benefit from diverse range of available unlabeled datasets. Learning from such a broad range of data not only yield a general-purpose solution, but it provides the opportunity of learning hidden and less studied patterns available in real-world data that may not be found in small annotated datasets which are usually captured under fully-controlled and restricted conditions in laboratories.

While self-supervised models can be used as a standalone model in some applications such as segmentation (in image and video \cite{caron2021dino}, or human activity or emotions \cite{deldari2020espresso}), object tracking \cite{valverde2021there}, anomaly and change point detection \cite{deldari2021tscp2}, they are mainly used as a \textit{pre-training} step for more complicated downstream tasks like classification.
Recent works in the field of Computer Vision (CV), such as SEER \cite{goyal2021self} and Dino \cite{caron2021dino}, and field of Natural Language Processing (NLP), such as BERT \cite{devlin2018bert} and XLM-R \cite{conneau2020xlm}, confirm self-supervise pre-training can yield noticeable improvement toward various downstream tasks compared to the fully-supervised trained models \cite{lecun2021dark}. However, learning from unlabeled or low-labelled sensor data is yet a challenging and open problem.
A model that is useful for ubiquitous computing applications must be able to deal with different types of sensor modalities, temporal resolutions, and other heterogeneity to extract general representations via disentangling noise from the underlying signal for downstream tasks, such as human activity/behaviour context recognition.

\subsection{Why Contrastive Learning?}
Amongst the self-supervised learning frameworks, Contrastive Learning (CL) has gained enormous attention due to its simplicity and effectiveness in training general-purpose encoders. The core principle behind contrastive methods is to learn features from pairs of similar samples (positives) and recognize them amongst negative distractors. Specifically, they do so through attracting representations of positive pairs closer to each other in the latent space and pushing dissimilar pairs apart with a contrastive objective. Furthermore, there are several advantages of the contrastive approach to other self-supervised deep learning models. Most prominently, as opposed to generative or auto-encoding techniques~\cite{goodfellow2014generative,zhang2017split}, it is highly efficient as it does not require a decoder and avoids input reconstruction.
In addition, in contrast to hand-crafted auxiliary tasks~\cite{saeed2021sense,gidaris2018unsupervised}, which may require extensive domain knowledge and hyper-parameter tuning, contrastive methods are task and domain-agnostic.

During the last two years there was a rapid progress in contrastive learning methods in the field of computer vision \cite{caron2021dino,chen2020simple,henaff2020data,wang2021clear,niizumi2021byol}, audio and speech processing \cite{saeed2021contrastive,niizumi2021byol,liu2021tera,wang2021contrastive,oord2018representation,xue2021exploring}, and natural language processing domain \cite{fang2020cert,devlin2018bert,oord2018representation,gao2021simcse,carlsson2021CTsemantic,wu2020clear,yan2021consert,giorgi2021declutr}. These models offers competitive performance compared to fully supervised methods. Contrastive approaches typically explored in the vision domain heavily depend on data augmentation (e.g., image rotation, image cropping, etc) to generate similar views of the input. Likewise, other approaches rely on implicit augmentation, where, similar data points are captured from the same modality but different viewpoints (e.g., camera angles)~\cite{Sermanet2017TCN} or utilizing signal decomposition methods (e.g., wavelet transform)~\cite{9141293}. This often requires prior knowledge of the problem domain which can be challenging in sensor applications. Another common issue that can arise with the hand-designed self-supervised auxiliary task is that the model may exploit shortcuts to quickly solve a task and learn features with no real value for the downstream tasks.

\subsection{Why Cross-modal Learning?}
Following the progress in the \textit{unimodal} contrastive learning models, \textit{cross-modal} and \textit{multimodal} contrastive models have been introduced and gained a lot of attention. 
Cross-modal or cross-view is a multimodal model learning approach where modalities are used as a supervisory signal for other modalities and the knowledge from one modality is distilled to another modality \cite{alwassel_2020_xdc}.  

Existing techniques in this area are mostly multimodal but not cross-modal, and they are trained in a supervised manner or regression-based approach (i.e. autoencoders). Existing multimodal deep learning models train modality-specific or modality-generic networks separately to extract modality-specific features, and then, combine modalities to consider the relationships between modalities. However, there are some weaknesses in these approaches, for example, late-fusion models do not consider inter-modality correlation, or in early fusion models, there is a tradeoff between inter and intra- modality features (general vs local). Moreover, they are trained for a specific downstream task and based on the available labelled dataset.

Considering the ability of human beings in learning from multiple sources of data with no or minimal supervision, cross-modal learning aims to learn from different modalities to capture a comprehensive and semantic understanding of the data. In this regard, several recent self-supervised deep learning models are proposed to learn from two or three modalities together, such as audio-video  \cite{ma2021active,morgado2021audio}, audio-text \cite{hsu2021hubert}, vision-text \cite{radford2021learning,huo2021wenlan,lei2021understanding}, audio-vision-text \cite{guzhov2021audioclip,akbari2021vatt,alayrac2020self}. 
Despite the undeniable need of considering data from many different sensor sources \cite{banos2021opportunistic} and the difficulty of annotation of fast-generating time-series data in ubiquitous applications, applications of self-supervised contrastive learning in \textit{cross-modal} time-series data is still not well-studied yet and they vastly focus on surrogate tasks \cite{saeed2021sense}, multimodal learning \cite{eldele2021tstcc,haresamudram2021contrastive}, or generative modeling~\cite{yao2018sensegan}. 

Furthermore, applying existing contrastive methods on multiple modalities requires pair-wise contrasting mechanisms between modalities \cite{tian2019contrastive,wang2021multimodal} which results in combinatorial computational complexity. Assuming the dot-product of sample pairs is the atomic operation of the contrastive objective function, the time complexity of contrasting each pair of modalities will be $\mathcal{O}(N^{2})$, where $N$ is the size of the input. Extending the number of modalities, $V \geq 2$ results in $\frac{V!}{(V-2)!}$ different combinations of modalities and complexity of $\mathcal{O}(V^2.N^{2})$, which is quadratic with respect to the number of modalities $V$ and batch size $N$.
On the other hand, the majority of prior contrastive models require a large number of negative pairs which means larger batch sizes, $N$, that can significantly increases the space and time complexity. On the other hand, unlike field of compute vision, in the field of wearable and ubiquitous sensors, larger batch sizes can potentially create a larger number of false negative pairs that can degrade performance \cite{deldari2021tscp2}.

In this work, we specifically focus on cross-modal learning. Regarding the fact that each source of data provides a complementary view or knowledge about the same event, each modality can be considered as a supervisory signal for the other modalities. Our work focuses on leveraging \textit{unlabeled} data from multiple devices to learn good quality features from the data, which can later be used to train different downstream tasks with limited labelled data. Furthermore, we show that our proposed model is more efficient in terms of computation and storage efficiency. 

\subsection{Contribution}


We propose \METHOD\ (Cross mOdality COntrastive leArning), an \textit{effective} and \textbf{\textit{ architecture}}- and \textbf{\textit{modality-agnostic}} contrastive learning framework by introducing a \textbf{\textit{cross-modal contrastive objective function}} to learn general-purpose representations from \textit{multimodal} time-series and provide \textit{data-efficient} solution for downstream tasks.

In \METHOD, we use the term `modality' in a broader sense that encompasses different types of data, sensors and input channels, and also can be easily reused in multiview problems. Unlike conventional contrastive methods, our approach does not require applying pre-defined augmentations for similar pair (or anchor-positive) generation. We instead leverage the different modalities of the input data to generate training examples, by treating samples from one modality as an anchor and temporally-aligned samples from other  modalities as the positive samples.
We hypothesize that in ubiquitous sensor processing, each sensor and modality provides a novel complementary view of the whole phenomenon. Hence we investigated our hypothesize across variety of sensor modalities including accelerometer, gyroscope,  electroencephalogram (EEG), electrooculogram (EOG), Electromyography (EMG), electrocardiogram (ECG), Electrodermal activity (EDA), etc. Although we focus on multimodal sensor data, \METHOD\ is applicable on multi-view problems as well.
The key contributions of our work can be summarized as follows:

\begin{itemize}
     \item \METHOD\ learns a compact joint representation between multiple modalities in a self-supervised manner. \METHOD\ is the first approach to apply contrastive learning to multiple modalities \textit{simultaneously} with lower time and memory complexity compared to previous methods that offered quadratic space and time complexity with respect to the number of data modes. Thus, our approach is far more efficient to learn self-supervised representations of high dimensional temporal datasets; where a large number of different sensing assets have been fused. Such datasets are widely found in applications of mobile and wearable sensing for augmented reality, smart home automation, and health domain that require contextual awareness.
     
     \item \METHOD~ is shown to be applicable over range of different sensor and devices for various applications (i.e. human activity recognition, sleep stage detection and emotion recognition) while existing rival methods only consider single type of sensors (mostly accelerometer data \cite{jain2022collossl,haresamudram2021contrastive}).
     
     \item As a cross-modality contrastive learning objective function, \METHOD\ can extract both cross- and intra-modality features and is shown to be able to outperform other supervised and self-supervised baselines models.

    \item  \METHOD\ is highly label efficient. It surpasses supervised baseline and other SOTA SSL models even using small subset of the labeled data to fine-tune the model on a downstream task. We also investigated the effectiveness of learnt representations toward downstream tasks in absence of fine-tuning.
      
\end{itemize}

\section{Related Work}
\label{sec:related_work}
In this section we review state-of-the-art contrastive learning models from the following points of view:

\begin{enumerate}
    \item Proposed contrastive-based objective function,
    \item Multimodal and cross modal contrastive-based models, and 
    \item Self-supervised and contrastive learning approaches in sensor data.
\end{enumerate}  

\begin{table}
\centering
\caption{Defintion of notations used in describing the problem and method}
\label{tab:notations}
\begin{tabular}{c|l}

\hline
\textbf{Notation} & \textbf{Meaning}                                                     \\ 
\hline
$V$                 & Number of modalities, views, or features         \\
$X_{v}$       & \textit{Set of input samples from modality $v$.}   \\
$x_{v}^{t}$       & \textit{The $t^{th}$ sample from modality $v$.}   \\
$z_{v}^{t}$       & \textit{The corresponding representation for the $t$th sample of modality $v$.} \\
$f(.)$            & \textit{The Encoder function}    \\
$S_{v,w}^{t,t'}$  & \textit{Similarity function between sample $x_{v}^{t}$ and $x_{w}^{t'}$}                            \\
$\tau$              & Temperature parameter (scale adjustment)  \\
$\mathcal{L}$       & Loss function  \\
\hline
\end{tabular}
\end{table}

\subsection{Contrastive Loss Functions}
\label{sec:loss_sec}
One branch of self-supervised learning involves contrastive learning-based approaches. The core of contrastive learning is to map raw data into a latent space in which positive samples are close to each other, whereas negative samples are pushed apart from each other.
Contrastive loss functions such as Triplet loss \cite{weinberger2009distance} are the most commonly used contrastive loss functions. However, they only consider one positive and one negative pair at a time which makes them converge slowly because while two negative samples are pushed apart they might be getting closer to other negative samples in the latent space. Multiple Negative Learning loss functions try to solve this problem by contrasting a sample against multiple negative samples simultaneously \cite{sohn2016improved}. These approaches, however, can be quite biased toward the sampling process and highly affected by the existence of false negatives in the training batch. Therefore, recently several objective functions have been introduced to implicitly perform Hard Negative Instance Mining \cite{wang2019ranked,sohn2016improved,duan2019deep,wu2017sampling} and debiasing \cite{NEURIPS2020_63c3ddcc,robinson2021hardcontrastive} in a self-supervised manner.

\begin{table*}
\centering
\caption{Summarizing existing self-supervised contrastive and non-contrastive learning loss functions. (CM shows if the objective function is proposed for multiple modality data.)}
\label{tab:loss_table}
\begin{tabularx}{\textwidth}{l|c|l|X}
\hline
\textbf{Method}  & \textbf{CM}*         & \textbf{Objective Function} &  \textbf{Details}   \\ 
\hline
 \begin{tabular}[c]{@{}l@{}} InfoNCE\\ (2018) \\ \cite{gut2010NCE,oord2018representation} \end{tabular}& - &  $- \sum_{x \in P}{log \frac{e^{f(x).f(x^{+})}}{e^{f(x).f(x^{+}))}+\sum_{x^{-}\in N}{e^{f(x).f(x^{-})}}}}$  & One shared encoder $f(.)$ for both of $view$ $v$ and $w$. \\
\hline

\begin{tabular}[c]{@{}l@{}}DCL \\ (2020) \\ \cite{NEURIPS2020_63c3ddcc} \end{tabular} & - &            \begin{tabular}[c]{@{}l@{}} $g = \max \{\frac{1}{\tau} ( \frac{1}{N} \sum{e^{f(x).f(x_{w}^{-})}} + e^{f(x).f(x_{w}^{+})} ) , \epsilon \}$ \\
$\mathcal{L}_{DCL} = - \sum_{t}{log \frac{e^{f(x_{v}^{t}).f(x_{w}^{+})}}{e^{f(x_{v}^{t}).f(x_{w}^{+}))}+ Ng}} $
\end{tabular} & Debiased Contrastive Learning for a positive sample and N negative samples across views $v$ and $w$ of single input dimension. The main goal is to reduce the effect of false negatives.\\
\hline

\begin{tabular}[c]{@{}l@{}}Hard-DCL \\ (2021) \\ \cite{robinson2021hardcontrastive} \end{tabular}& - &
\begin{tabular}[c]{@{}l@{}} $ w = \frac{\beta e^{f(x,x^{-})}}{\sum_{x^{-}}{e^{f(x,x^{-})}}}$ \\
$g = \max \{\frac{1}{\tau} ( \frac{1}{N} \sum{w * e^{f(x).f(x^{-})}} + e^{f(x).f(x^{+})} ) , \epsilon \}$ \\
$\mathcal{L}_{Hard\_DCL} = - \sum_{t}{log \frac{e^{f(x^{t}).f(x^{+})}}{e^{f(x^{t}).f(x^{+}))}+ Ng}} $
\end{tabular} 
& Improves Debiased Contrastive Learning by increasing weight of hard negative samples.  \\
\hline
\begin{tabular}[c]{@{}l@{}} Barlow-\\Twins \\ (2021)\\ \cite{zbontar2021barlow} \end{tabular}& -  &   
\begin{tabular}[c]{@{}l@{}}$C_{ij} = \frac{\sum_{t}{f(x_{i}).f(x_{j})}}{ \sqrt{\sum_{t}{f(x_{i}}^{2}} \cdot \sqrt{\sum_{t}{f(x_{i})}^{2}}}$\\ $\mathcal{L}_{\Barlow} = \sum_{i}{(1-C_{ii})^{2}} + \lambda \sum_{i}{\sum_{j \ne i}{C_{ij}^{2}}}$ \end{tabular} & Non-contrastive objective function that only considers positive pairs. It aims is to increase the similarity of paired representations while reducing the redundancy.   \\ 
\hline

\begin{tabular}[c]{@{}l@{}} $CMC$ \\ (2019)\cite{tian2019contrastive} \\ (2021) \cite{wang2021multimodal} \end{tabular} &Yes & 
$\mathcal{L}_{CMC} = \sum_{v \in V}{\sum_{w \in V, w \ne v}{\mathcal{L}_{NCE}(v,w)}} $&
\textit{CMC} calculates the contrastive InfoNCE loss function over every pairs of views. \\
\hline
\METHOD\ &Yes & $\mathcal{L}_{\METHOD} =  \sum_{t}{\mathcal{L}_{C}^{t}} + \lambda \sum_{v \in V}{\mathcal{L}_{D}^{v}}$ & 
A combination of the cross-modality correlation($\mathcal{L}_{C}$) and intra-modality discriminator ($\mathcal{L}_{D}$).Explained in Method Section.\\
\hline
\end{tabularx}
\end{table*}

To highlight how \METHOD\ is distinguished from other existing objective functions and provide a full comparison, we summarize recent state-of-the-art methods (that we used as baselines), their proposed objective functions in Table \ref{tab:loss_table}. It is worth mentioning that the table does not aim to cover all works that have been done on self-supervised contrastive learning (e.g., it does not cover works with new network architecture but similar loss function), but only those with novel objective functions are included. In Table \ref{tab:notations} we have explained the notation we used in the following sections to describe our objective function and also other state-of-the-art methods.

\subsection{Multimodal Contrastive Learning} 
Contrastive learning methods mostly have been proposed for single modality data across a variety of applications including computer vision~\cite{he2020momentum,caron2020unsupervised,chen2020simple,niizumi2021byol}, audio processing~\cite{saeed2021contrastive,niizumi2021byol,xue2021exploring}, natural language processing ~\cite{fang2020cert,giorgi2021declutr}, sensor data analytics~\cite{9141293,saeed2021sense,franceschi2019unsupervised,cheng2020subject,kiyasseh2020clocs,tonekaboni2021unsupervised,eldele2021tstcc,deldari2021tscp2}. Recently \cite{deldari2022beyond} has reviewed existing self-supervised learning approaches for multimodal and temporal data including contrastive learning models.
Although there are many works on learning from multimodal data, they conventionally relied on a single modality learning and processed each mode of data individually. On the other hand, recent works are focusing on the combination of two or more different modalities to use the co-occurrence of events in multiple modalities as a supervisory signal to train modality-specific/generic networks together simultaneously \cite{arandjelovic2017look}. For example text data can perfectly supervise image classification \cite{radford2021learning} or audio classification \cite{guzhov2021audioclip}. 

Existing multimodal contrastive learning techniques are mostly applied to jointly learn embeddings of text and vision (e.g., images and videos) for purposes such as extracting captions for the images and videos \cite{yuan2021multimodal,radford2021learning},  contrasting text against audio \cite{hsu2021hubert} for speech recognition, or vision against audio \cite{wang2021multimodal}. However, expanding the number of views will cause the time complexity to be in combinatorial order of available views/modalities. Table \ref{tab:multimodal} summarizes recent cross-modal SSRL models.

\begin{table}[htbp]
  \centering
  \caption{Recent SOTA contrastive models on cross modal learning.}
    \begin{tabular}{l|l|l}
    \toprule
    \textbf{Name} & \textbf{Loss Function} & \textbf{Modalities} \\
    \midrule
    TNC(2017) \cite{sermanet2017time} & Triplet loss and regression & Two camera views (vision)\\
    MMV(2020) \cite{alayrac2020self} &   MIL-NCE,Info-NCE & 3 modalities(Video, Audio, Text) \\
     $CMC$(2020) \cite{tian2019contrastive} &  Info-NCE & Multiple views/augmentation (Vision) \\
    XDC(2021) \cite{alwassel_2020_xdc} & Clustering& 2 modalities (video, audio)\\
    CLIP(2021) \cite{radford2021learning}&   Info-NCE & 2 modalities (Image, Text) \\
    VATT(2021) \cite{akbari2021vatt} &  MIL-NCE,Info-NCE & 3 modalities(Video, Audio, Text) \\
    \bottomrule
    \end{tabular}%
    \label{tab:multimodal}%
\end{table}%

\subsection{Contrastive Models for Wearable Sensor Data }

Self-supervised models benefit from abundant amount of available unlabelled datasets to learn compact but informative features (representation) of the underlying data. The gained knowledge in the form of pre-trained models and extracted features will be employed through few-shot learning and transfer learning applications where only a limited amount of labelled data is available.

Existing Self-Supervised Representation Learning (SSRL) models proposed for time-series data (e.g. wearable and environment sensor data), usually employ predefined tasks to pre-train the models without requiring external label or annotated data. In \cite{saeed2019multi}, a shared encoder is trained to learn multiple pretext tasks including multiple signal transformations (e.g. noise, signal scaling, signal rotation, signal permutation, etc.). The main goal is to learn useful describing features of input data that can detect the type of transformation applied to the original signal. 
Afterwards, the trained encoders will be used or fine-tuned to learn the downstream task objective function. Similarly, authors of \cite{saeed2021sense} and \cite{sarkar2020self} pre-train the backbone network based on various signal transformation tasks as pretext tasks, and employ the pre-trained model to learn from limited labelled datasets. 
Instead of signal transformation, authors in \cite{haresamudram2020masked} trained an encoder-decoder model as a pretext task to reconstruct the masked parts of the original input. 

Apart from the pretext-based models, recent SOTA self-supervised representation learning methods tried to pre-train backbone encoders using a variant of the contrastive objective function. Following the effectiveness of contrastive learning models in computer vision and natural language processing, many research have been also customized and extended the similar CL framework on the ubiquitous sensor and time-series data
 \cite{deldari2021tscp2,tonekaboni2021unsupervised,eldele2021tstcc,tang2021selfhar,saeed2019multi,sheng2020weakly}. Similar to pretext models, the contrastive-based pre-trained encoders are then employed to improve the performance of the downstream task. Contrastive models have been applied to sensor data in various applications including but not limited to emotion recognition \cite{mohsenvand2020contrastive}, 
sleep stage classification \cite{mohsenvand2020contrastive,eldele2021tstcc},
medical purposes \cite{tonekaboni2021unsupervised},
anomaly detection\cite{mohsenvand2020contrastive},
seizure recognition \cite{eldele2021tstcc}, 
human activity recognition (HAR) \cite{eldele2021tstcc,tonekaboni2021unsupervised,haresamudram2021contrastive,khaertdinov2021contrastive},
changepoint detection (CPD) \cite{deldari2021tscp2}. 

One of the most important factors that impact the performance of contrastive learning is the choice of positive and negative pairs \cite{ge2021robust}. Existing CL approaches for time-series data can be divided mainly into the following categories depending on how they generate the positive pairs:
\begin{itemize}
    \item Augmented pairs: These models aim to learn contextual information by applying different signal transformations (e.g. signal inversion, noise addition, temporal cutout, frequency cutout, sensor shuffling \cite{cheng2020subject}) and contrasting the representations of each transformed signal. The main idea is that regardless of the transformation head applied to original data, the learnt representation should be similar and share the same context. However, the selection of the suitable set of transformation/augmentation is an important task as we need to make sure the transformed signals can preserve the information required by the downstream task \cite{tang2020exploring,tian2020makes}. A detailed review of common augmentation methods for time-series data can be found in \cite{wen2021augmentation}.
    
    \item Temporally related pairs: These methods extract positive and negative pairs based on the temporality of data, so the temporally adjacent sequences of input time-series form a positive pair. The main goal is to learn representations from history (past sequences) that can predict the representations of the future time windows \cite{deldari2021tscp2,tonekaboni2021unsupervised,haresamudram2021contrastive,banville2021uncovering}. 
    
    \item Temporal-Contextual pairs: 
    These methods generate positive pairs by applying both signal transformation and temporal prediction techniques. For example, \cite{eldele2021tstcc} applies two sets of weak and strong augmentation (signal transformation) across the original sequence of data. While the proposed contrastive solution forces the model to learn similar representations for both of the weak and strong augmentations of the same instance (augmented pairs), the learnt representations should be able to predict the future representations of the other augmented signal (temporally related pairs). Similarly, in ~\cite{kiyasseh2020clocs}, a contrastive learning-based approach is proposed for cardiac signals,  to extract patient-specific signal representations by employing both temporal and contextual contrasting.
\end{itemize}

Table \ref{tab:relatedwork} summarizes existing contrastive-based  techniques on sensor data.
While existing contrastive learning-based methods have shown promising results in learning the representations of temporal sensor data, it remains a challenge to design a \textit{cross-modal} contrastive learning framework with temporal sensor data. Authors of \cite{franceschi2019unsupervised} investigate representation learning on multivariate time-series data \cite{franceschi2019unsupervised}, However, their method depends on triplet losse and require explicit mining of negative pairs to train the model, which is quite challenging in practice. On the other hand, prior works on self-supervision for sensors mainly focuses on learning representations from one or two modalities, does not consider the multimodal (or multi-sensor) nature of the data \cite{tonekaboni2021unsupervised}. They ignored cross-modality information that can improve the model's predictive performance through leveraging inter-modality relationships.

\begin{table*}[htbp]
  \centering
  \caption{Recent SOTA contrastive models on sensor data. CM states if the model is proposed for cross modal learning or not.}
    \begin{tabularx}{\textwidth}{l|c|l|X|X}
    \toprule
    \textbf{Name} & \textbf{CM} & \textbf{Loss Function} & \textbf{Category}& \textbf{Application} \\
    \midrule
    CSSHAR(2021)\cite{khaertdinov2021contrastive} & No & NT-Xent\cite{chen2020simple} & Augmentation & HAR\\
    -(2020)\cite{cheng2020subject} & No & Info-NCE & Augmentation & Biosignals (EEG, ECG)\\
    TCN(2020) \cite{tonekaboni2021unsupervised}& No    & Triplet Loss & temporal & General time-series \\
    TS-CP2(2020) \cite{deldari2021tscp2} & No    & Info-NCE & temporal & General time-series \\
        
    CPC(HAR)(2021) \cite{haresamudram2021contrastive} &No & Info-NCE & temporal & HAR\\
    
    CPC(EEG)(2021)\cite{banville2021uncovering}& No & Info-NCE & temporal & EEG signals\\
    
    CLOCS(2020) \cite{kiyasseh2020clocs}& No & Info-NCE & Augmentation + temporal & Medical (ECG signals)\\
    
    TS-TCC(2021) \cite{eldele2021tstcc}& No    & Info-NCE & Augmentation + temporal & \\

    ColloSSL(2022)\cite{jain2022collossl}& Yes & Info-NCE & multi-device cross view& HAR\\

    \METHOD(ours) & Yes   & proposed & multimodal cross view &General time-series  \\
    \bottomrule
    \end{tabularx}%
    \label{tab:relatedwork}%
\end{table*}%

Instead of generating augmented or temporal positive samples, we proposed to employ synchronous data segments across available sensors and modalities to create positive samples due to the fact that each modality captures the same phenomenon but from a different perspective. The most closely tied to our work is the very recent and promising work on human activity recognition, ColloSSL \cite{jain2022collossl}. ColloSSL is a collaborative device self-supervised learning model that considers data from multiple accelerometers and gyroscope sensors. To extract positive samples, ColloSSL selects the most suitable sensor(s) (device) based on the Maximum Mean Discrepancy (MMD) metric. The main assumption here is that data captured by relevant sensors necessarily have a similar distribution. In the case of having various types of sensors (such as datasets we studied in this work) this assumption is not valid anymore. In \METHOD\, instead of considering the sensors with the most similar distribution, we hypothesize each modality can contribute towards a complementary understanding of the same underlying factors. In other words, data from all modalities can effectively help to form the relevant representation in the latent space as they are related to the same label/class of event regardless of the difference in their distribution. Heterogeneity between devices is also confirmed to be beneficial and can help towards generalizability of the model in federated learning applications that deal with multiple devices and sensors \cite{cho2021device}.
Hence, instead of a single encoder, we employ modality-specific encoders which also makes our model robust to missing modalities during inference and training of the downstream task.  

Recently, \cite{qian2022makes} provided a practical comparison using several state-of-the-art contrastive learning frameworks. Although the employed baselines are proposed for another domain (mainly computer vision), they applied them across wearable sensor data for human activity recognition tasks and studied the effect of different modules such as augmentation techniques and the backbone network. However, this work also overlooked the main gap in applying contrastive learning for multimodal sensor data such as a large number of modalities, which leads to input data with various characteristics and distributions, and also the limited number of classes compared to other domains. While the usage of ubiquitous sensor data is rapidly increasing and the annotation of these complex fast-growing data is infeasible, still, there is a gap in applying contrastive methods that are computationally applicable to the high number of modalities. 


\section{METHOD}
\label{sec:method}
In this section, we, first, formally describe the problem and then explain our proposed technique to learn from multiple views of unlabeled data based on contrastive learning. We also discuss how the proposed approach differs from previous self-supervised contrastive learning models. Table \ref{tab:notations} explains the notation we used in the following sections to describe our proposed objective function.

\subsection{Problem Definition}
Given a multimodal sensor data $\{X_{1}, X_{2},\cdots,X_{V}\}$, $V$ is the number of sensors or sources of data and $X_{v}=\{x_{v}^{1}, \cdots, x_{v}^{T}\}$ shows $T$ set of observations of modality $v$. We consider different sensor modalities to be equivalent to the different views, and hence, we can use the terms interchangeably. The $t$th sample is a list of observations from all modalities and is defined as $x^{t}=[x_{1}^{t}, \cdots, x_{V}^{t}]$. We attempt to map each modality $v$ of sample $x_{v}^{t}$ to its representation vector $z_{v}^{t}$, through its modality specific encoding function $f_{v}(.)$, where $z_{i}^{t}, t \in T, i \in V$ belongs to a shared space $Z$. Our goal is to learn a compact informative representation of data that is shared between all modalities and can effectively represent the class and state of the system.  

The main components of any contrastive learning models are 1) positive and negative sampling, and 2) the objective functions. In the next sections we explain our proposed approach for each module.

\subsection{\METHOD: Positive and Negative Pairs Sampling}
\begin{figure*}[]
    \begin{minipage}[t]{.64\textwidth}
        \vspace{20pt}
        \subfigure[]{
        \label{fig:cocoa_model}
        \includegraphics[width=0.9\linewidth]{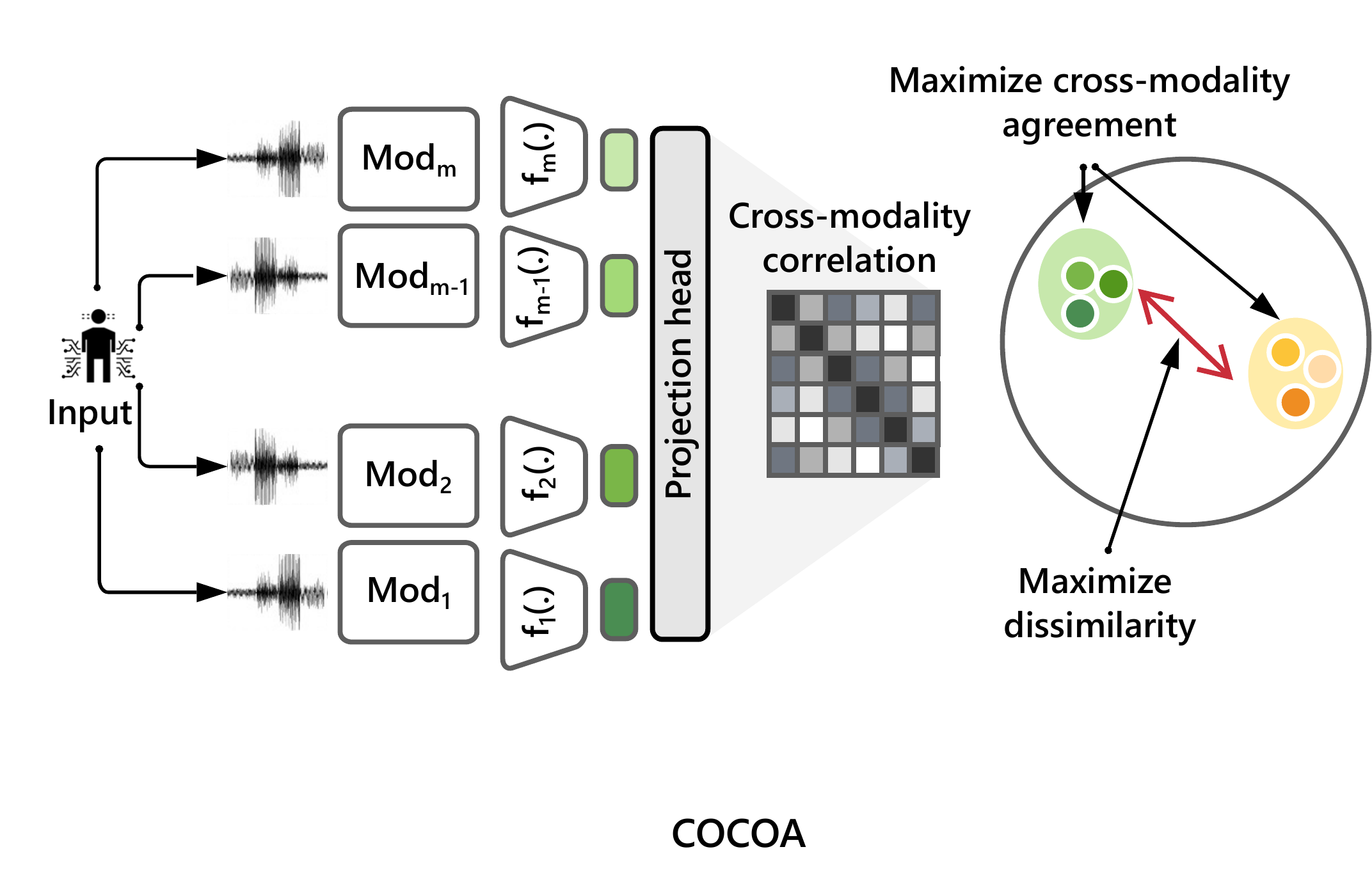}}
    \end{minipage}
    \hfill
    \vline
    \begin{minipage}[t]{.34\textwidth}
    \vspace{0pt}
         \subfigure[]{
         \label{fig:contrastive_model}
         \includegraphics[ height=1in]{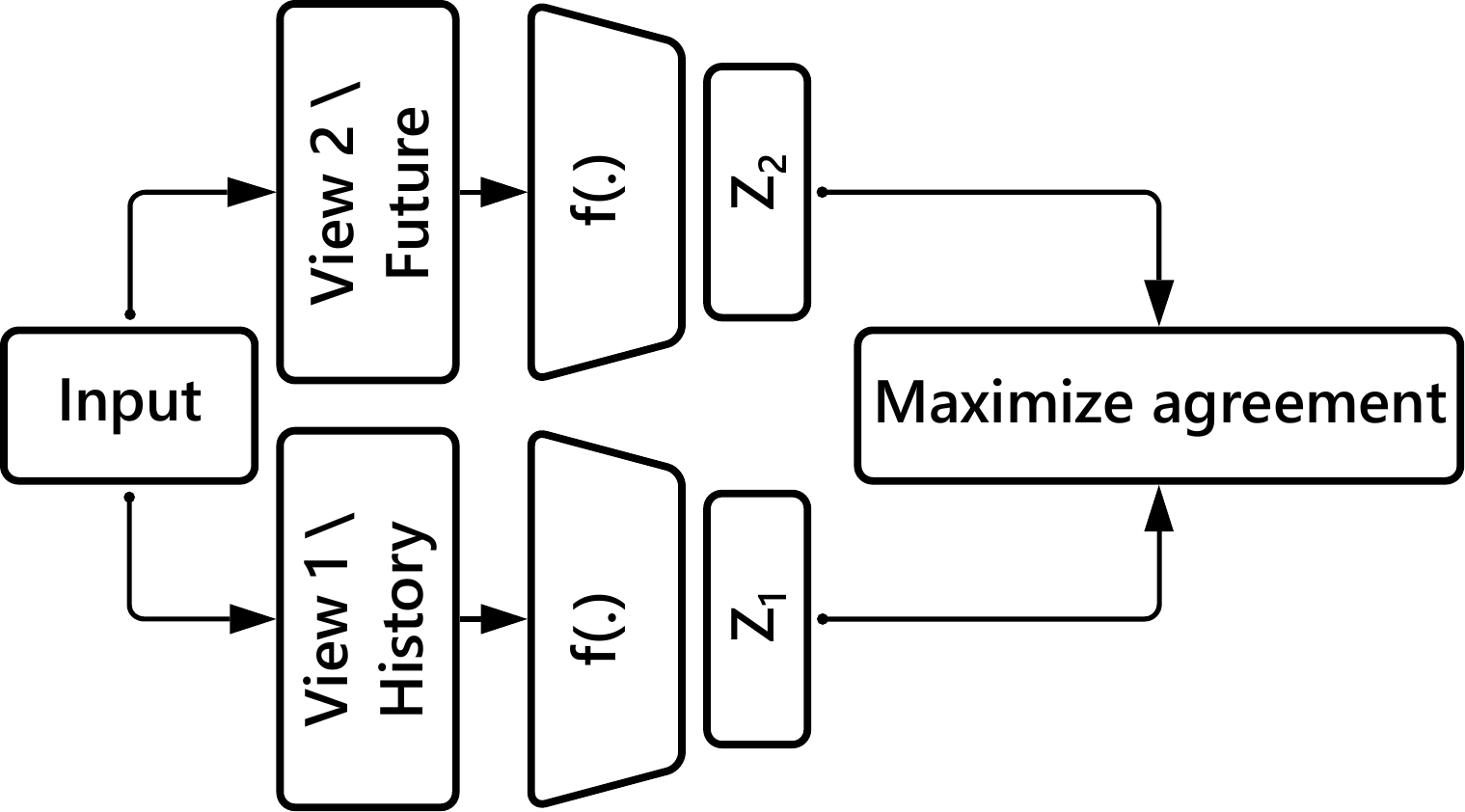}} \\
         \subfigure[]{
         \label{fig:multiview_model}
         \includegraphics[height=1.4in]{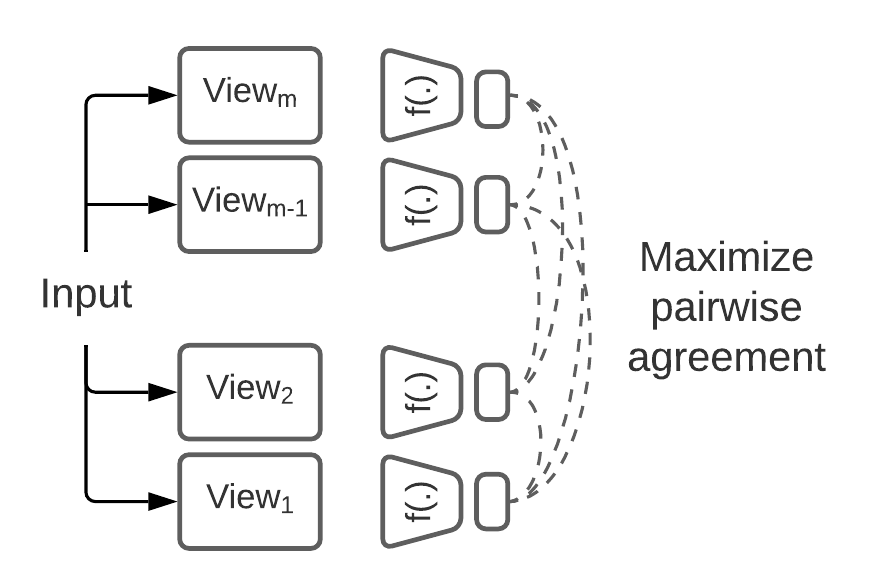}
         }
    \end{minipage} 
   \caption{Illustration of \METHOD\ architecture as a cross-modality contrastive learning model [\ref{fig:cocoa_model}] vs. Two pair contrastive learning [\ref{fig:contrastive_model}] and pair-wise multi-view contrastive models [\ref{fig:multiview_model}]}.  
   \label{fig:arch}
\end{figure*}

Recent self-supervised contrastive learning methods are commonly based on comparing representation vectors of \textit{two} distorted version of the same input instance \cite{zbontar2021barlow,chen2020simple,henaff2020data,saeed2021contrastive,he2020momentum,niizumi2021byol}, known as augmented views, or based on the history and future time intervals of a data instances  \cite{oord2018representation,deldari2021tscp2, tonekaboni2021unsupervised}. 
In \METHOD\, $views$ are not limited to different augmentation or distortion, of the same input data, but $views$ can be defined as various modalities or sensors as well.
Figure \ref{fig:cocoa_model} shows \METHOD\ extracts a joint representation from modality-specific encoders by optimizing the model to maximize the agreement between modalities at the same time step  (\textit{positive pairs}) and minimize the agreement between samples in each batch that are sampled from random and temporally distant time-steps (\textit{negative pairs}) within the shared space $Z$.

A standard contrastive learning method trains the model to learn a similar representation between paired \textit{views} of each sample (Figure \ref{fig:contrastive_model}). In these models, a \textit{view} refers to an augmented version of a sample, or a proceeding time interval of a sample within the latent space . Multi-view and multimodality models (including our proposed approach), however, expand the number of contrastive branches to consider several views of each sample to learn from a more diverse set of views of the data. Existing multimodal models, as is shown in Figure \ref{fig:multiview_model}, still contrasts single pairs of views at each time and aggregates the proposed loss function across all pair-wise combinations of the views \cite{tian2019contrastive,wang2021multimodal,akbari2021vatt}. These frameworks suffer from slow convergence, partially because they are only employing a single pair of views at each time. Therefore, while they try to make the representation of single pairs of views more similar, this may push other pairs of views further away from each other. Moreover, as we explained in the Introduction Section, there is computational and memory cost to expand such approaches to higher dimensional data.

Considering the calculation of dot-product between each pair of embedding vectors (representation) as the atomic function that requires one unit of CPU computation and 1 unit of memory, the cost for each training cycle of models with full-contrasting objective functions (Figure \ref{fig:multiview_model}) is quadratic with respect to the number of views $\mathcal{O}(V^{2}N^{2})$ where $V$ is the number of modalities and $N$ is the batch size. While due to pruned set of positive and negative pairs, \METHOD~'s complexity will be in order of $\mathcal{O}(V^{2}N+VN^{2})$ (including the cost for cross-modal positive pair comparison ($O(V^{2}N)$) and inter-sample negative pair comparison ($O(VN^2)$)). Since \METHOD~ tends to perform better with smaller batch sizes (as it is shown in Section \ref{sec:sens_anly}), the cost function will reduce to $O(V^{2})$ for higher number of modalities. 

\begin{figure}
    \centering
    \includegraphics[width=0.8\linewidth]{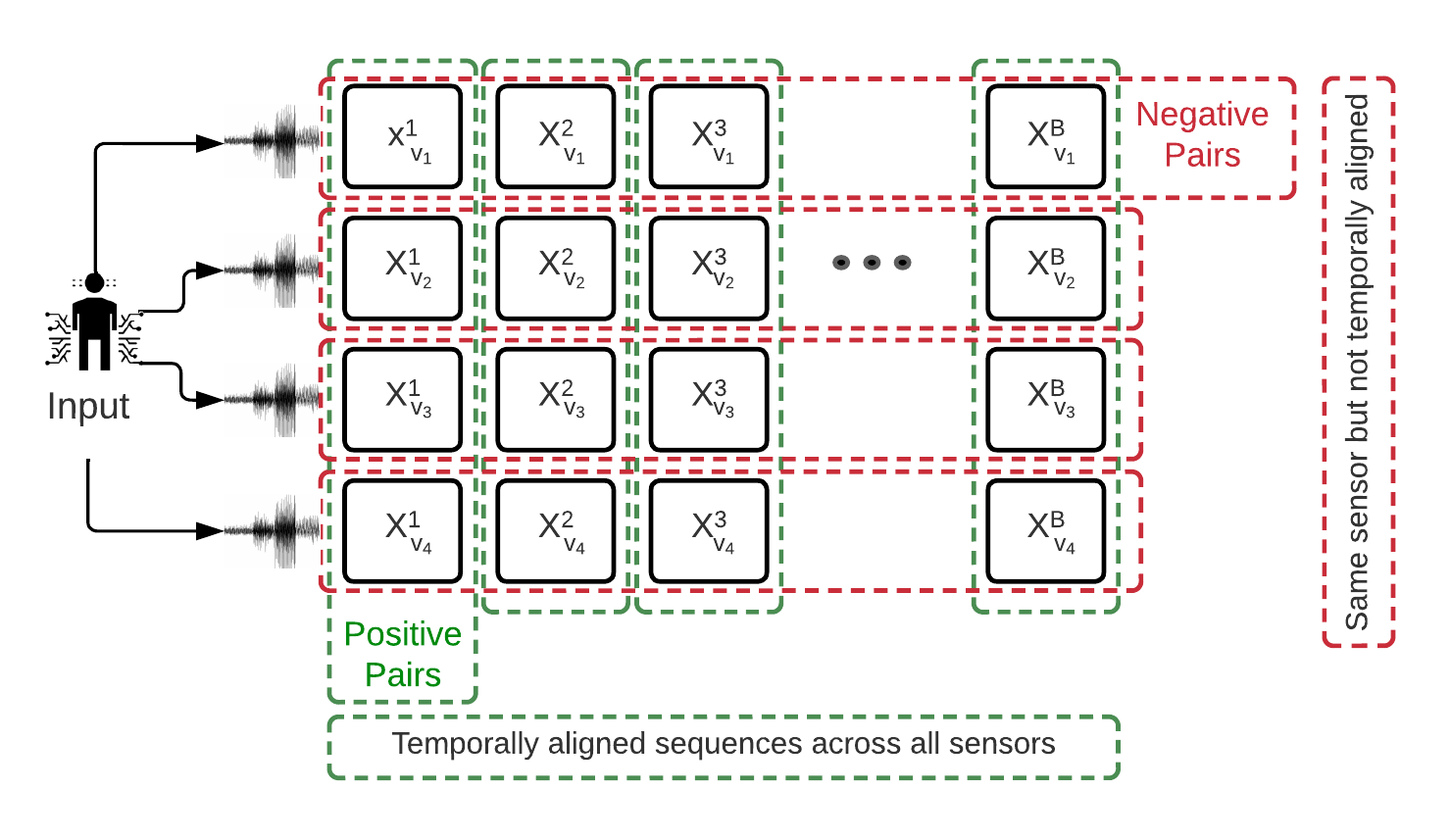}
    \caption{Illustration of \METHOD's positive and negative sampling method. ($B$ represents the batch size)}
    \label{fig:sampling}
\end{figure}

Figure \ref{fig:sampling} visualise our proposed sampling approach. Timely aligned readings across all available sensors are considered as positive pairs (inter-modal), while negative pairs are extracted from temporally distant samples within each individual sensor (intra-modal).
The main goal is maximizing the agreement between the positive pairs' embeddings $(f(x_{v}^{t}),f(z_{w}^{t})), v,w \in V$ and minimizing the similarity between embeddings of negative pairs $(f(z_{v}^{t}),f(z_{v}^{t'}))$.

\subsection{\METHOD: Objective Function}

In \cite{tian2019contrastive}, the authors proposed Contrastive Multiview Coding, \textit{CMC}, objective function upon Noise Contrastive Estimation \cite{gut2010NCE}. 
The Full view formulation of \textit{CMC} objective function, contrasts all pair-wise combinations of existing views ($L_{F} = \sum_{v \ne w}{L(X_{v},X_{w})}$). Thus, the full view has a combinatorial cost with respect to its number of views, which means it fails to effectively scale to a large number of views. In contrast, we propose a simple and novel contrastive loss function to consider all available views at the same time with linear space and time complexity. The objective function consists of following vital parts: 1) \textbf{\textit{Cross-modality correlation}}, and the 2) \textbf{\textit{Discriminator}}. 

\subsubsection{Cross-modality correlation}
\label{sec:corr}
The first part of the objective function calculates cross-correlation between latent encodings of different
modalities of an individual sample, $x^{t}$, as follows:

\begin{equation}
 \label{eq:positive_term}
  \mathcal{L}_{C}^{t} = \sum_{w \neq v}{e^{(\frac{1-S_{v,w}^{t,t}}{\tau})}}\ %
\end{equation} 

\begin{equation}
 \label{eq:similarity_term}
  S_{v,w}^{t,t'} = \frac{z_{v}^{t} \cdot z_{w}^{t'}}{\|z_{v}^{t}\| \cdot \|z_{w}^{t'}\| } \ %
\end{equation} 
where $\tau$ is adjusting hyper-parameter (also called temperature parameter \cite{chen2020simple}) and $S_{v,w}^{t,t'}$ is normalized dot-product of representations of the modalities $v$ and $w$ for samples $t$ and $t'$, respectively. We believe considering synchronous samples from different modalities, make it a \textit{harder positive pair} compared to temporally adjacent samples from the same modality \cite{haresamudram2021contrastive} because of difference in distribution of data across different source of input.

\subsubsection{Discriminator} 
\label{sec:disc}
Along with maximizing inter-modality agreement, the representation of irrelevant samples need to be discriminated from each other. Therefore, the objective function should also be able to minimize the similarity between latent encoding of different samples within a batch. To simplify the discriminatory function, we consider random and temporally far samples within each individual view (modality) $v$, ($x_{v}^{t}$ and $x_{v}^{t'}$), as negative pairs and minimize the agreement between them: 

\begin{equation}
 \label{eq:negative_term}
  \mathcal{L}_{D}^{v} = \frac{1}{T} \sum_{t,t'}^{T}{e^{(\frac{S_{v,v}^{t,t'}}{\tau})}}\ %
\end{equation} 

Given the samples of each individual modality are probably more similar rather than samples from different modalities, they will form \textit{harder negative pairs}. It is shown that considering harder negative pairs makes the contrastive objective function learns more effectively by extracting the most distinguishing features \cite{kalantidis2020hard}.

By combining the cross-view correlation ($\mathcal{L}_{C}$) and discriminator ($\mathcal{L}_{D}$) loss functions, the loss function of \METHOD\ is formulated as follows:

\begin{equation}
 \label{eq:loss_func}
  \mathcal{L}_{\METHOD} =  \sum_{t}{\mathcal{L}_{C}^{t}} + \lambda \sum_{v \in V}{\mathcal{L}_{D}^{v}}%
\end{equation} 
where $\lambda$ is a constant weight parameter for adjusting the importance of the discriminator term (negative pairs). 

\METHOD\ loss function has similarities with the recently introduced self-supervised objection function, \Barlow\ \cite{zbontar2021barlow}. \Barlow\ measures the cross-correlation between embedding vectors of the same sample generated by \textit{two identical} networks fed with \textit{two augmented versions} of the original sample. The purpose is to learn a representation where the cross correlation matrix between its pairs of embedding vectors are pushed close to the identity matrix. The diagonal elements of the cross-correlation matrix, representing the corresponding dimensions of the embedding vector pairs, are pushed to be maximally correlated. Whilst the redundancy between the off-diagonal elements of the cross-correlation matrix, representing different embedding vector dimensions are minimised. The \Barlow\ loss function operates upon each sample independently without contrasting the batch samples against each other. Therefore, there are no negative pairs in the loss function. However, an implicit contrast is applied by the batch normalization layer in the implemented network \cite{yuandong2021nonegative,tsai2021note}.

\subsection{\METHOD: Contributions}
Finally, we can highlight the contributions of \METHOD~ in 4-folds as follows: 

\begin{enumerate}
\item \textit{Cross-modal.} Existing multimodal sensor approaches are mostly based on the fusion of the modalities before, during or at the very late layers of the training model (mostly done by concatenating the raw/embedded data). However, our work is considered cross-modal learning which not only uses the raw data as a supervisory signal but also uses mutual information across modalities as the main supervisory signal. We evaluated the effectiveness of cross-modal compared to multimodal learning in section \ref{sec:xu}.
    Moreover, \METHOD\ expands the cross-correlation similarity measure to \textit{more than two} modalities so that modality-specific networks can be leveraged across a variety of input ubiquitous sensor data instead of image data only.     
    \item \textit{Implicit sampling of Hard positives, hard negatives.} As we explained in Section \ref{sec:corr} and \ref{sec:disc}, \METHOD~ leads to reducing the possibility of the number of fake negatives in self-supervised training which is a challenging issues in contrastive learning techniques \cite{jain2022collossl,deldari2021tscp2,qian2022makes}. The effectiveness of the proposed technique is compared against other approaches that includes all possible positive and negative samples \cite{tian2019contrastive}. We have shown even with a small fraction of comparisons (limited positive and negative pairs), we can achieve higher performance compared to the fully supervised model. 
    
    \item \textit{Computation and memory efficiency.} In the Discriminator part of the loss function, \METHOD\ contrasts the negative pair samples of each mini-batch. We reduce the complexity by removing the unnecessary comparisons and contrasting among modalities across different time intervals. We hypothesize that this contrasting criterion can be achieved implicitly by maximizing the cross-modality correlation and maximizing the dissimilarity across negative samples (discriminator). 

    \end{enumerate}

We showed the effectiveness of our less complicated model through extensive experiments in the next section. To highlight how \METHOD\ distinguished from other existing models and provide a full comparison, we summarize recent state-of-the-art contrastive objective functions in Table \ref{tab:loss_table}.

\subsection{Architecture and Implementation Details}
Figure \ref{fig:cocoa_model} illustrates the high-level architecture of \METHOD. In this model, we propose to use modality-specific encoders to map the heterogenous input data into a shared latent space. The encoders are followed by a single modality-specific dense layer projection head and a shared dense layer to fuse the modalities during the training process. 
For the encoders $f_{m}(.)$, we use a temporal convolutional network that contains a stack of $3$ convolution layers having kernel size of $10$, $8$, and $4$ with filter sizes of $24$, $48$, and $20$ followed by layer normalization. However, \METHOD\ is agnostic to the underlying architecture of encoders. Each encoder is followed by a specific and a shared dense layer as projection and modality fusion head, respectively. We used Adam optimizer with a learning rate of $0.001$ and an early stopping condition on the loss function. We conducted experiments over a different range of batch sizes for the purpose of sensitivity analysis.

The main purpose of the experiments is to investigate and compare the effectiveness of the proposed objective function, \METHOD\, in learning more informative representations from multiple sources of data, against other SOTA contrastive objective functions. Since the proposed objective functions are encoder-agnostic, and to provide a fair comparison across existing objective functions, we used the same underlying architecture (as mentioned above) for all baselines. Therefore, the baselines only represent the contrastive objective function proposed in their corresponding paper but not the network architecture. 
The implementation is based on Python3.8 and Tensorflow2 \footnote{\url{https://www.python.org/downloads/release/python-380/}, \url{https://www.tensorflow.org/}}. To make \METHOD\ reproducible, all the code and experiments are available on the project's webpage\footnote{\url{https://github.com/cruiseresearchgroup/COCOA}}.


\section{Experiments}
\label{sec:xu}
In this section, we first introduce and describe the properties of the datasets and the SOTA baselines used in evaluating our model. Next, we present the result of our evaluation experiments to 
\begin{itemize}
    \item Examine the contribution of considering additional modalities simultaneously.
    
    \item Investigate the performance of \METHOD\ in learning informative representations compared to its supervised version and other self-supervised baselines.
    
    \item Study data-efficiency of \METHOD\ against other self-supervised and supervised baseline, regarding the amount of available labeled data.
    
    \item Compare sensitivity analysis toward range batch sizes.
    
    \item And finally, illustrate and study the distribution of extracted representations.
\end{itemize} 

And finally, in \ref{sec:discussion} we discuss challenges, findings, and future directions in this domain.

\subsection{Datasets}
\label{sec:datasets}
To evaluate our approach, we use human activity recognition, sleep stage detection, and emotion analysis datasets as these tasks usually come with a variety of sensor modalities. We evaluate \METHOD\ on five different well-known public datasets: UCIHAR \cite{anguita2013ucihar}, SLEEPEDF \cite{kemp2000analysis,goldberger2000physiobank}, Opportunity \cite{roggen2010collecting}, PAMAP$2$ \cite{reiss2012pamap2}, and WESAD \cite{wesad}. 

UCIHAR dataset \footnote{\url{https://archive.ics.uci.edu/ml/datasets/human+activity+recognition+using+smartphones}}
is a human activity recognition dataset that contains recordings from 30 volunteers who carried out 6 activities including \textit{walking, walking upstairs, walking downstairs, sitting, standing,} and \textit{laying}. Activities are recorded by a smartphone device mounted on volunteer's waist.


PhysioNet Sleep-EDF
dataset \footnote{\url{https://physionet.org/content/sleep-edf/1.0.0/}} 
consists of 61 polysomnograms (PSGs) recordings from 20 subject for two whole nights and contains recordings of 2 electroencephalogram (EEG), electrooculography (EOG), and electromyography (EMG) sampled at the rate of $100$Hz. The signals are annotated by sleep experts into $8$ stages of sleep including \textit{Wake (W), Rapid Eye Movement (REM), N$1$, N$2$, N$3$, N$4$, Movement} and \textit{Unknown (not scored)}. We applied standard pre-processing as proposed in \cite{supratak2017deepsleepnet} to merge N$3$ and N$4$ stages into a single class following American Academy of Sleep Medicine standards and removed the unscored and movement segments. 

PAMAP2 \footnote{\url{https://archive.ics.uci.edu/ml/datasets/pamap2+physical+activity+monitoring}}, Physical Activity Monitoring dataset, contains data of 18 different physical activities performed by 9 subjects wearing 3 inertial measurement units and a heart rate monitor. In this set of experiments we only used 3 accelerometor sensor data and 9 activities including \textit{sitting, standing, walking, running, cycling, nordic-walking, ascending stairs, descending staris}, and \textit{jumping} which are sampled at rate of 100HZ.

WESAD \footnote{\url{https://ubicomp.eti.uni-siegen.de/home/datasets/icmi18/}}
, Multimodal Dataset for Wearable Stress and Affect Detection, is a well-known stress and emotional affect dataset that includes recordings from vraiety of sensors including Acceleromenter, ECG, EMG, EDA, Temperature, BVP and respiration data collected from a RespiBAN Professional \footnote{\url{http://www.biosignalsplux.com/en/respiban-professional}} 
and an Empatica E4 \footnote{\url{http://www.empatica.com/research/e4/}} 
placed around the subject’s chest and subject's non-dominant hand, respectively. We used Accelerometer, ECG, EMG and EDA signals from E4 wristband and removed samples labelled as \textit{non-defined} and considered samples with \textit{baseline, stress, amusement}, and \textit{meditation} labels. 

Opportunity \footnote{\url{http://opportunity-project.eu/}},
this dataset consists of data collected from IMU sensor from 4 participants performing activities of daily living with 17 on-body sensor devices. For our evaluation, we followed the same settings used by recent work \cite{jain2022collossl}.

Table \ref{tab:dataset-table} provides more detail about the number of sensors, subjects, classes and sizes of each dataset. 
To obtain data for training, we use a sliding window technique to extract window with $50\%$ overlaps. For UCIHAR dataset, as suggested by \cite{saeed2019multi} and \cite{tang2021selfhar}, we used windows with 400 samples. Similarly, for PAMAP2 and Opportunity datasets we followed the existing works \cite{jain2022collossl} and extracted windows of $5.12$ and $2$  seconds (512 and 60 smaples), respectively. For WESAD and SLEEPEDF datasets, we considered the same window size suggested by the dataset publisher and as it is used in other research works \cite{saeed2021sense,deldari2021tscp2} and extracted samples with length of $1000$ ($10$ seconds) and $3000$ ($30$ second), respectively.

\begin{table*}[]
\caption{Characteristics of the datasets used in our evaluation. The number in \{\} represents the number of classes.}
\label{tab:dataset-table}
\centering
\begin{tabularx}{\textwidth}{llcccX}
\toprule
\textbf{Application} & \textbf{Dataset} & \textbf{Modality}   &\textbf{Subjects} & \textbf{Size} & \textbf{Classes}\\ \midrule

\multicolumn{1}{l}{\multirow{3}{*}{ \begin{tabular}[l]{@{}l@{}}Human Activity \\  Recognition \end{tabular}}} & \begin{tabular}[l]{@{}c@{}}UCIHAR\\ \end{tabular}           & \begin{tabular}[c]{@{}c@{}} Acc, Gyro\end{tabular}   & 30 & 2.5K & \textit{walking, walking up/downstairs, sitting, laying, standing} \{6\}    \\ \cline{2-6} 

\multicolumn{1}{c}{}     & \begin{tabular}[c]{@{}c@{}}PAMAP2\\  \end{tabular}     & \begin{tabular}[c]{@{}c@{}}3xAcc  \\ (arm, chest, ankle) \end{tabular}         & 9 &11K   &        \textit{sitting, standing, walking, running, cycling, nordic-walking, ascending/descending stairs}, rope-jumping \{9\}     \\ \cline{2-6}

\multicolumn{1}{c}{}     & \begin{tabular}[c]{@{}c@{}}Opportunity\\  \end{tabular}     & \begin{tabular}[c]{@{}c@{}}5xAcc  \\ (back, left L-arm, right\\ U-arm, left/right shoe) \end{tabular}         & 4 &23K   &        \textit{standing, walking, sitting, lying}   \\ \hline

\begin{tabular}[l]{@{}l@{}}Sleep Stage\\ Detection \end{tabular} & \begin{tabular}[c]{@{}c@{}}SLEEPEDF\\ \end{tabular}          & \begin{tabular}[c]{@{}c@{}}2xEEG, EOG, EMG\end{tabular}            &20 & 55K  &  \textit{Awake, Rapid Eye Movement, N$1$, N$2$-N$3$}, and \textit{N}$4$    \{5\}         \\ \hline

\begin{tabular}[l]{@{}l@{}}Emotion\\ Recognition \end{tabular}& \begin{tabular}[c]{@{}c@{}}WESAD\\  \end{tabular}     & \begin{tabular}[c]{@{}c@{}}Acc, ECG, EMG, EDA\end{tabular}   & 15   & 21K    &        \textit{baseline, stress, amusement}, and \textit{meditation} \{4\}      \\

\bottomrule
\end{tabularx}
\end{table*}

\subsection{Experimental Setup}
\subsubsection{Baselines}
We compared the proposed objective function, \METHOD\, in improving the performance of the downstream task (i.e., classification) against two supervised and eight self-supervised baselines to provide a comprehensive evaluation in three different aspects:

\begin{enumerate}
    \item Effectiveness of SSL pre-trained model compared to fully supervised techniques. For the sake of completeness and providing a supervised reference, we compare the performance of \METHOD\ to the supervised baseline publicly accepted model, DeepConvLSTM \cite{ordonez2016deep} which we call it DeepCL, and a network with the same architecture as \METHOD\ which we call it ``sup.'' for short. However, the supervised baselines are trained from scratch in two different setups: end-to-end (E2E) and with frozen randomly initialized encoders (Fixed). For DeepCL we reused the implementation from \cite{hoelzemann2020digging}\footnote{\url{https://github.com/ahoelzemann/deepconvlstm_keras}}. 
    \item Effectiveness of the proposed contrastive objective function: we compared \METHOD~ with other contrastive frameworks including InfoNCE \cite{oord2018representation}, DCL \cite{NEURIPS2020_63c3ddcc}, Hard-DCL \cite{robinson2021hardcontrastive}, \Barlow~ \cite{zbontar2021barlow}, $CMC$ \cite{tian2019contrastive} and $CPC$ \cite{haresamudram2021contrastive}. Details of each objective function is explained in Section \ref{sec:loss_sec} and Table \ref{tab:loss_table}.
    
    \item Effectiveness of cross-modal learning: we compared \METHOD~ against three different multimodal frameworks: 1) Contrastive Predicting Coding ($CPC$) \cite{haresamudram2021contrastive}.  2,3) Two AutoEncoder baselines \cite{baldi12autoencoder}. Following the suggested baseline comparison framework by \cite{jain2022collossl}, we employed AutoEncoder-Single (AE-S) with a single shared encoder and AutoEncoder-Multi ($AE-M$) with multiple modality-specific encoders, to extract representation from input data. We used a decoder to convert the representations to original input data. After training the encoder+decoder model based on minimizing the Mean Squared Error (MSE), we discard the decoder module and use the trained encoder for the downstream task. 
\end{enumerate}

We have discussed the details of these SSL loss functions in the previous sections in more detail.
It is important to note that we only compared the proposed objective functions as they are the focus of this work. To avoid the effect of complex backbone and different network capacity \cite{qian2022makes}, and make a fair comparison among all baselines, the underlying network architecture (encoder) is assumed to be the same across all baselines and experiments. Hence, we employed a simple encoder network to measure only the capability of the objective functions. However, \METHOD~ is an encoder-agnostic objective function and can be easily integrated with more complicated and powerful networks such as transformers. 

For self-supervised pre-training of encoders, we discard all the labels. Following the evaluation framework in \cite{tian2019contrastive,jain2022collossl}, a linear classifier was used to evaluate the effectiveness of learned representations by employing a softmax layer on top of the pre-trained encoders.
Furthermore, to provide a thorough analysis, based on the evaluation process suggested in \cite{henaff2020data}, we examined the quality of representations in two different setups that will be explained later including (1) Fixed and (2) Fine-tuned encoders. 

\subsubsection{Evaluation process and metrics.} We randomly split all datasets into $80$\%-$20$\% for training and test sets with no overlap of users for WESAD and SLEEPEDF dataset and no overlap of temporally close samples for HAR datasets. We further divided the training set to create a validation set of size $10$\%, which is used for hyper-parameter tuning and early stopping.
We repeated each training experiment five times and reported the average macro F1-score (unweighted mean of F1-scores over all classes) as a performance metric.

\subsubsection{Hyper-parameter tuning}
We conduct grid search over the hyper-parameters including temperature scales, $\tau=[ 0.1, 0.5, 1]$, and range of different batch sizes (between $8$ and $1024$), and reported the best result for each method. We used initial learning rate equal to $0.001$ for both self-supervised and classification training. We employed Adam optimizer \cite{kingma2014adam} and early stopping mechanism to stop the training after five epochs without progress. 
All models are trained for 100 epochs.

\subsection {\METHOD: Cross-modal Representations}
As we discussed in Section \ref{sec:method}, by providing a general contrastive objective function, \METHOD\ can learn high quality representations in a self-supervised manner from multiple views (modalities). We hypothesize that the different modalities/views not only share information but also each can provide distinct aspects of the task at hand as well. 
Since most of existing SSL contrastive baselines are proposed for only uni- or dual-modal data and they are not able to contrast more than two views of the data, we conducted two sets of experiments: 

\begin{enumerate}
    \item $M=2$: To make the other SOTA SSL cost functions (i.e. DCL, HardDCL, NCE, Barlow) comparable to \METHOD, we trained the models based on only two sensors at a time. The experiments are repeated for different two-pair combination of available modalities for each dataset and the final result is the average of all experiments. For example in case of SLEEPEDF dataset with four types of sensor, we investigated the performance of \METHOD~ and all other baselines across all six different sensor combinations, separately.
    \item $M>2$: In this set of experiments we considered all of the available modalities per dataset to investigate the effectiveness of including additional modalities.
\end{enumerate}
Table \ref{tab:all_table} compares the effectiveness of \METHOD\ and the other baselines in various setups including:
\begin{itemize}
    \item \textbf{Fixed}: The modality-specific encoders are pre-trained based on the contrastive objective function in a self-supervised manner; they are fixed during the classification training step to evaluate the quality of the SSL encoders without fine-tuning. In the supervised case, the encoders are randomly initialized and they are kept fixed for classification training. The intuition behind using fixed random encoders is to provide a \textit{lower bound} to evaluate the effectiveness of the extracted representations from the SSL models.
    
      \item \textbf{E2E}: The modality-specific encoders are pre-trained based on the contrastive objective function in a self-supervised manner and then fine-tuned according to the downstream task loss function. In supervised cases, the encoders and classifiers are trained jointly.
    
   \end{itemize}


Table \ref{tab:all_table} reports the average F-score achieved by each method across the best batch sizes (batch sizes varied from $8$ to $1024$ samples) along with the standard deviation. Comparing multiple modalities ($M>2$) against dual modality ($M=2$) confirms our hypothesisat contrasting more modalities leads to higher performance. Thus, extra modalities can play an additional supervisory role within self-supervised learning. In multiple modality experiments ($M>2$), we exclude UCIHAR as it contains only $2$ types of sensors.

\begin{table*}[]
\caption{Evaluating the performance of \METHOD, against supervised and self-supervised baselines with respect to the F1-score metric. We considered two cases: 1) E2E: Fine-tuned pre-trained encoders in an end-to-end manner, and 2) Fixed: we fixed the pre-trained encoder and only trained the classifier head. Reported values are averaged over 5 randomized experiments and the numbers in the parenthesis shows the standard deviation.}
\label{tab:all_table}
\centering
\npdecimalsign{.}
\nprounddigits{2}
\small
\footnotesize
\begin{tabular}{l|cc|cc|cc|cc|cc}
\toprule
\textbf{Datasets}   & \multicolumn{2}{c|}{\textbf{UCIHAR}}            & \multicolumn{2}{c|}{\textbf{PAMAP2}}            & \multicolumn{2}{c|}{\textbf{WESAD}}              & \multicolumn{2}{c}{\textbf{SLEEPEDF}}           &
\multicolumn{2}{c}{\textbf{OPPORTUNITY}}\\ \hline

Baseline                    & \textbf{E2E}           & \textbf{Fixed}         & \textbf{E2E}           & \textbf{Fixed}         & \textbf{E2E}           & \textbf{Fixed}          & \textbf{E2E}            & \textbf{Fixed}        &
\textbf{E2E}            & \textbf{Fixed}  \\ \hline

\multicolumn{11}{l|}{\textbf{M=2} }   \\ \hline

Sup.&91.9(1.4) & 71.3(7.2) & 89.3(2.9) & 26.7(8.3) & 55.6(8.8) & 22.8(2.4) & 50.4(8.9) & 25(2.4) & 88.5(3.1) & 66.8(3.5) \\ \hline

Barlow. & 93.6(1.5) & 57.5(3.4) & 91.1(3.4) & 18.9(3.6) & 62.1(15.9) & 30.1(8.7) & 51(8.7) & 24.9(3.6) & 90.3(3.4) & 23.8(8) \\

DCL          & 94.7(1) & 89.8(1.7) & 97.2(0.7) & 90.5(2.7) & 90.7(4.7) & 61.6(7.4) & 60.7(4.2) & 53.7(2.0)  & 93.4(0.9) & 85.6(2.3) \\

HardDCL      & 94.8(0.9) & 91(3.5) & 97.2(0.6) & 90.4(4.1) & 90.2(4.6) & 61.9(8.4) & 60.7(4.4) & 55.2(5.2) & 93.3(2.3) & 85(1) \\

NCE          & \textbf{95.9(1.3)} & \textbf{91.4(4.4)} & \textbf{97.6(1.2)} & 90.9(2.9) & 90.5(4) & 62.8(7.6) & 60.5(4.2) & 55.5(0.9) & 93.5(0.6) & \textbf{85.7(0.8)} \\

\textbf{\METHOD}        & 95.8(1.3) & 90.9(2) & 97.4(0.4) & \textbf{91(2.9)} & \textbf{91.5(2.9)} & \textbf{63.1(8.6)} & \textbf{62.2(1.9)} & \textbf{57.4(1.3)} & \textbf{93.9(1.3)} & \textbf{85.7(1.2)} \\
\bottomrule

\multicolumn{11}{l|}{\textbf{M$>$2} }    \\ \hline

Sup.        & -(-)   & -(-)   & 93.9(1.2) & 41.9(5.7) & 75.7(6.1) & 25.7(2.8) & 57.1(1) & 33.3(2.5) & 93.2(1.4) & 79.8(2.5) \\ 
 \textit{$DeepCL$}  & 93.6(2.2)  & -(-) & 95.6(0.6)& -(-) & 71.9(3.8) & -(-) & 67.4(1.7) & -(-) & 93.2 (1.7) & -(-) \\ \hline

\textit{$AE-S$}  & 92.7(1.3) & 52.6(4.8) & 96.8(0.7) & 9.9(8.9) & 72.1(15.4) & 37(4.1) & 52(4.3) & 26.5(1.3) & 94.7(0.3) & 48.7(9.2) \\

 \textit{$AE-M$} &95.1(1.4) & 89.5(2.7) & 96.9(0.8) & 33.8(15.9) & 94.1(2.5) & 48.6(2.9) & 60.3(2.1) & 53(0.9) & 96.3(0.5) & 82.5(6.6) \\
 \textit{$CPC$}  &  93.1(2) & 57.7(16.9) & 96.4(1.1) & 46.7(5.6) & 51.2(4.4) & 44.9(1.8) & 58(1.8) & 27.9(1.3) & 94.8(0.7) & 39.2(11.2) \\
 \textit{$CMC$}  & -(-)  & -(-)  & 98.3(0.1) & 94.9(0.3) & 96.4(0.2) & 80(8.2) & 66.9(1.2) & \textbf{62.8(1)} & 95.9(0.6) & 92.4(0.4) \\

\textbf{\METHOD}  & \textbf{95.8(1.3)} & 90.9(2) & \textbf{98.5(0.4)} & \textbf{96.3(1.4)} & \textbf{97.6(0.5)} & \textbf{87.6(1.7)} & \textbf{67.8(1.5)} & \textbf{62.8(2.3)} & \textbf{97.1(0.6)} & \textbf{93.3(0.5)} \\ \bottomrule

\end{tabular}
\npnoround
\end{table*}

\subsubsection{Fixed Encoders}
\label{fix_enc}
Comparing the classification result of baselines with a Fixed-encoder against the lower bound (the supervised baseline with a random encoder) and also the fully supervised baselines clearly shows that the self-supervised pre-trained encoders are playing a key role in extracting useful information and learning generic features from the input data, even though they do not have access to class labels.
Since the pre-trained encoders are frozen during classification training, the learned representations are directly used as input for the classification module. We observe that the representations are highly useful given the fact that a linear classifier can discriminate underlying classes with reasonably good recognition performance. In all studied datasets, \METHOD\ achieves the best result in both cases of dual and multimodal experiments except in the UCIHAR dataset where \METHOD\ achieved a highly competitive result with NCE. However, \METHOD\ is based on NCE with some improvements which make it suitable for multimodal data. 
Considering all available sensors ($M > 2$) in the Fixed setup, \METHOD\ achieves higher performance with all other baselines across all datasets. Moreover, \METHOD\ ($M>2$) improves the performance of dual-modal version ($M=2$) by $5.2$\%, $24.5$\%, $5.4$\% and $7.6$\%
in terms of average F-score across PAMAP2, WESAD, SLEEPEDF and Opportunity datasets, respectively. 
 \METHOD\, also, provides higher F-score compared to  $AE-S$ ($+51.2$\%), $AE-M$ ($+24.7$\%), $CPC$ ($+41.9$\%), and $CMC$ ($+2.5$\%), and the supervised baseline with random encoder (Fixed Rand.) ($+39.8$\%). According to the experiments, representations from pre-trained encoders of $AE-S$ and $CPC$ methods are less successful in classifying the samples. We believe this is due to the fact that both of these approaches concatenate all the modalities and do not consider inter-modal information which is the main challenge in this area. Secondly, although CPC(HAR) is successful in predicting future frames' representations based on the history (context), we noticed the GRU-based autoregressive module used in the $CPC$ architecture does not perform well for large sequences (large history) similar to window sizes we used in this work (see Section \ref{sec:datasets}).  
 
Although it is shown that \Barlow can effectively learn invariant features by bringing closer the representations of augmented views \cite{zbontar2021barlow}, as the results show, \Barlow does not perform very well on our datasets. However, in our case, we assumed each modality as an implicit augmented view, which may not be applicable for the \Barlow's proposed objective function and backpropagation mechanism through the modality branches.


\subsubsection{Fine-tuned Encoders.}
\label{f_t_enc}
According to Table \ref{tab:all_table} \METHOD\ fine-tuned encoder outperforms all other baselines across all of the datasets. This finding confirms the ability of \METHOD~ in learning effective representations, even compared to a fully supervised baseline. To be specific, fine-tuned \METHOD\ (in both cases of $M=2$ and $M>2$) outperform their corresponding fully-supervised baselines by $8.1$\% and $4.5$\% upon the PAMAP$2$, $35.8$\% and $21.9$\% across the WESAD, $10.6$\% and $11.7$\% across the SLEEPEDF and $5.4$\% and $3.8$\% in Opportunity dataset. 

Considering even only two modalities, the results confirm that the Fine-tuned \METHOD\ outperforms each of the other baselines, including the supervised baseline, for the WESAD, SLEEPEDF, and Opportunity datasets and provides the second-best result after the NCE across UCIHAR and PAMAP2. \METHOD\ not only outperforms SOTA models but also can consider a large number of modalities to further improve the quality of representations. 
To provide a fair comparison and discard the possible effect of batch size, we reported the average over the best batch size for each method. Surprisingly, all of the methods were superior across smaller batch sizes (e.g. 8 or 16) except $CPC$ which requires larger batch sizes (eg. 256 or 512) (We discuss this in more detail in Section \ref{sec:sens_anly}).

Comparing multimodal approaches ($M>2$) \METHOD\ provides higher F-score compared to  AutoEncoder-single ($+9.6$\%), AutoEncoder-Multiple ($+2.8$\%), $CPC$ ($+12.6$\%), and $CMC$ ($+0.8$\%), and also the fully supervised baseline ($+10.2$\%). Comparing crossmodal techniques (ie. \METHOD\ and $CMC$) with multimodal approaches (ie. autoencoder-based and CPC) confirms our hypothesis that crossmodal techniques can provide more insights into the underlying structure and correlation between modalities.

It is worth mentioning that ColloSSL \cite{jain2022collossl} also customised contrastive learning for human activity recognition tasks. ColloSSL \cite{jain2022collossl}, mainly proposes an end-to-end solution to tackle the cross-modality problem and includes device selection, pair sampling, and contrastive objective function strategies. ColloSSL proposed an extension of the Info-NCE loss function which can handle multiple positive pairs. However, we did not consider it as a baseline because the framework considers only one sensor as an anchor. Deciding on the anchor sensor not only requires background knowledge but also is not deterministic and can be changed per class, because each type of class (i.e. activity in this domain) may have a different set of dominant sensors or devices. Hence background knowledge is needed to find the most dominant device for each class of data and train different models separately. In contrast, \METHOD\ takes into account all the available modalities together and simultaneously. 

\begin{figure*}[ht]
\centering
    \subfigure[]{\includegraphics[width=.24\linewidth]{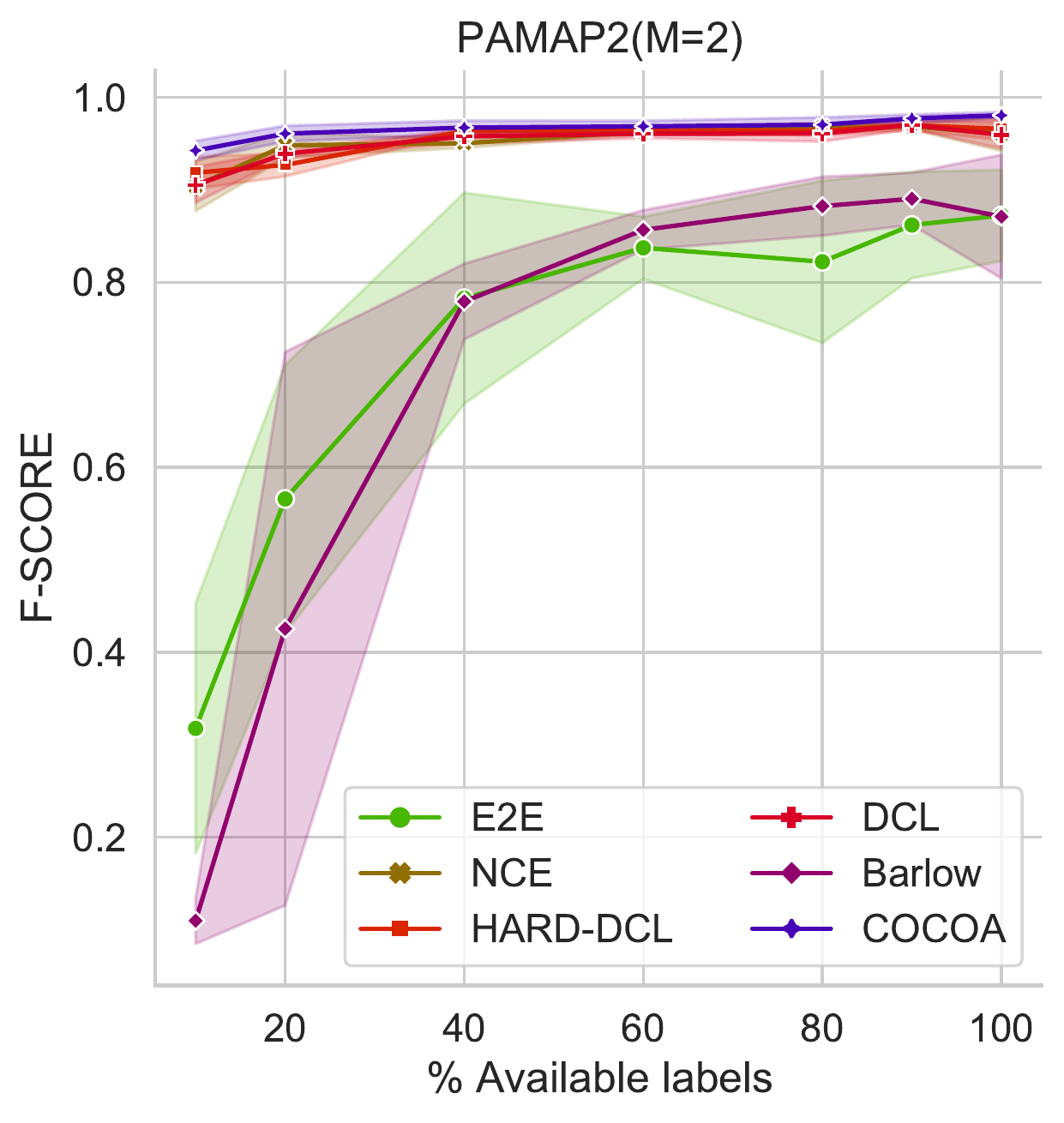}\label{fig:pamap_lb}}
    \subfigure[]{\includegraphics[width=.24\linewidth]{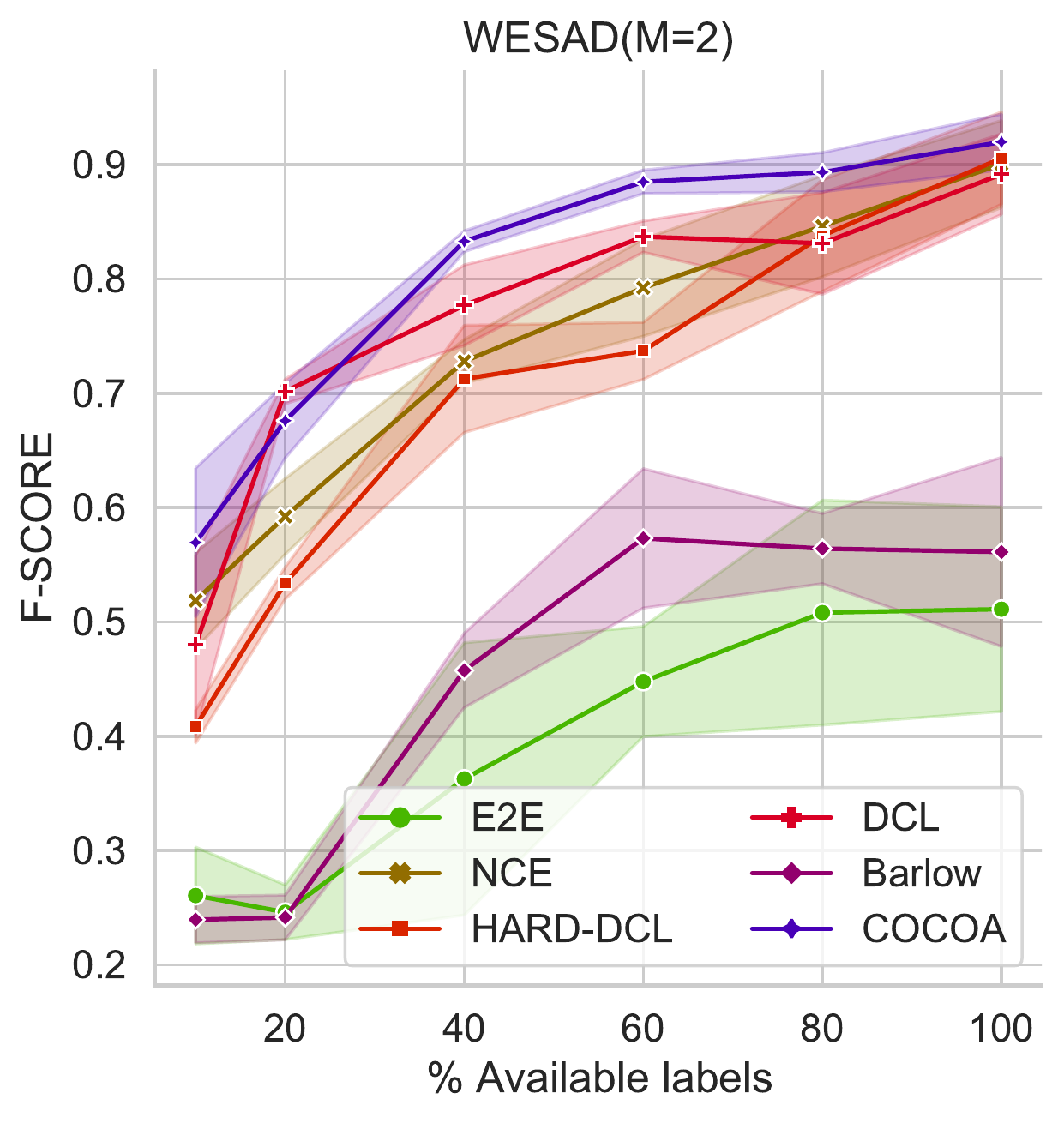}\label{fig:wesad_lb}}
     \subfigure[]{\includegraphics[width=.24\linewidth]{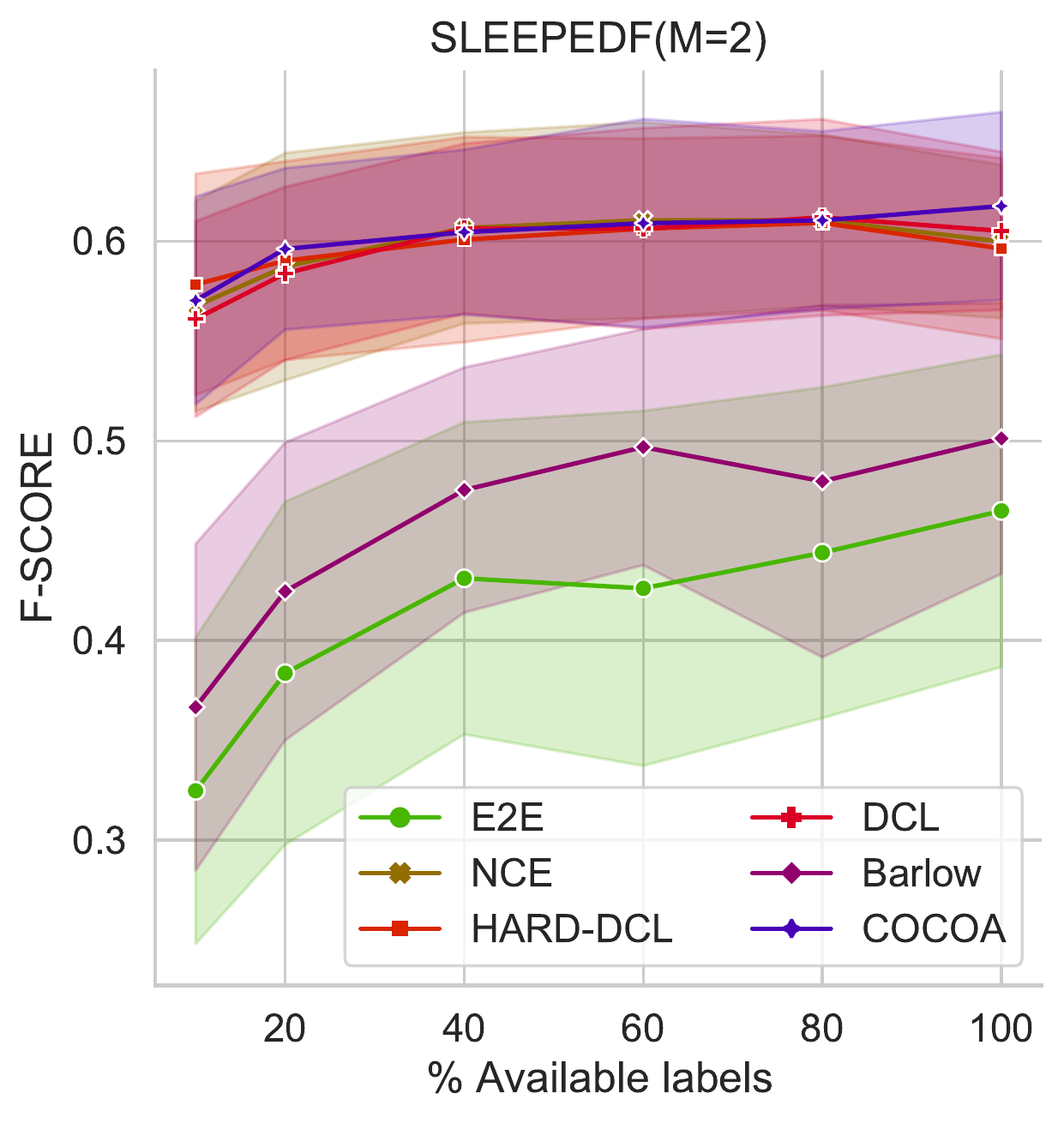}\label{fig:sleep_lb}}
    \subfigure[]{\includegraphics[width=.24\linewidth]{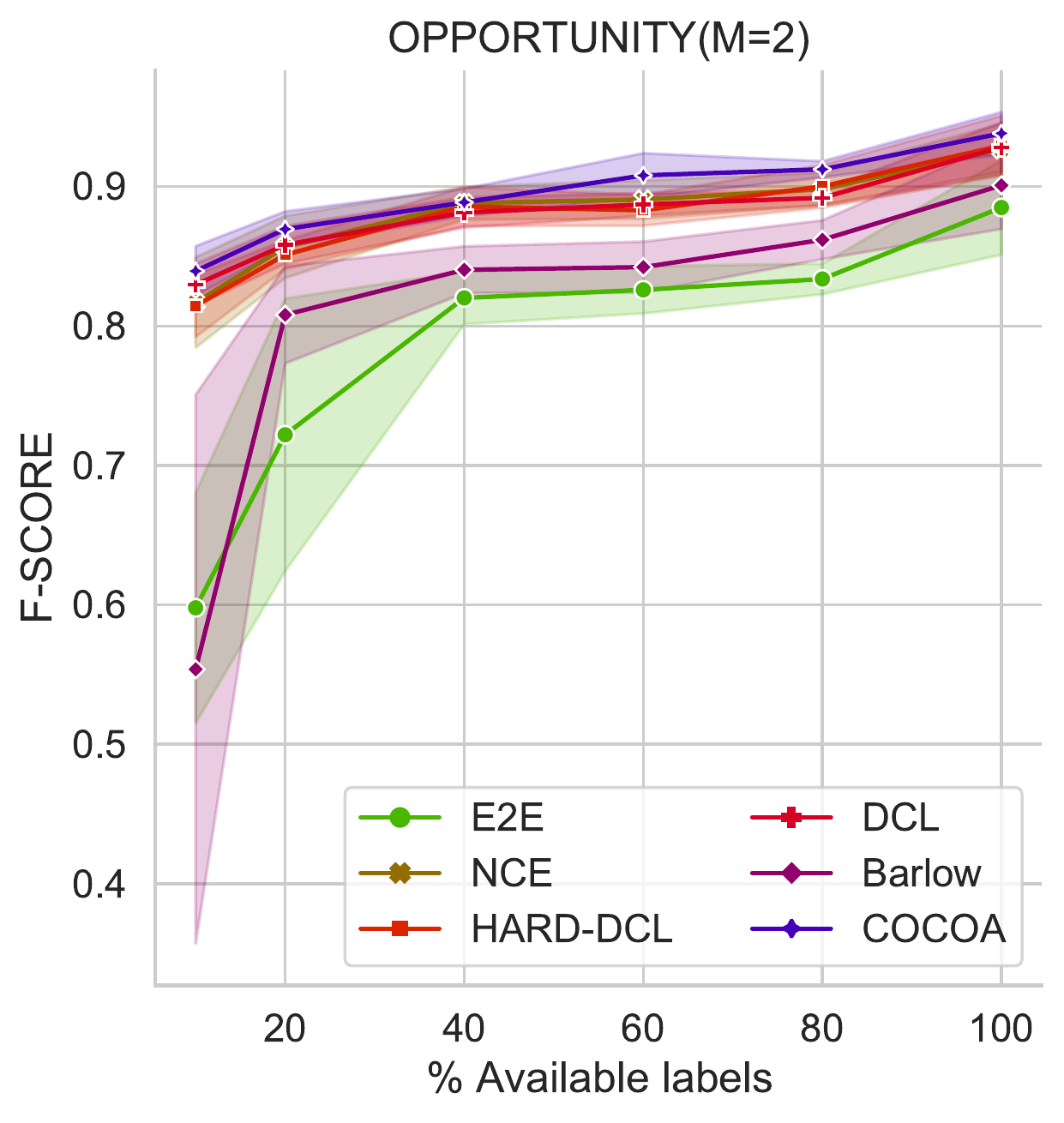}\label{fig:opp_lb}}
    \\
    \subfigure[]{\includegraphics[width=.24\linewidth]{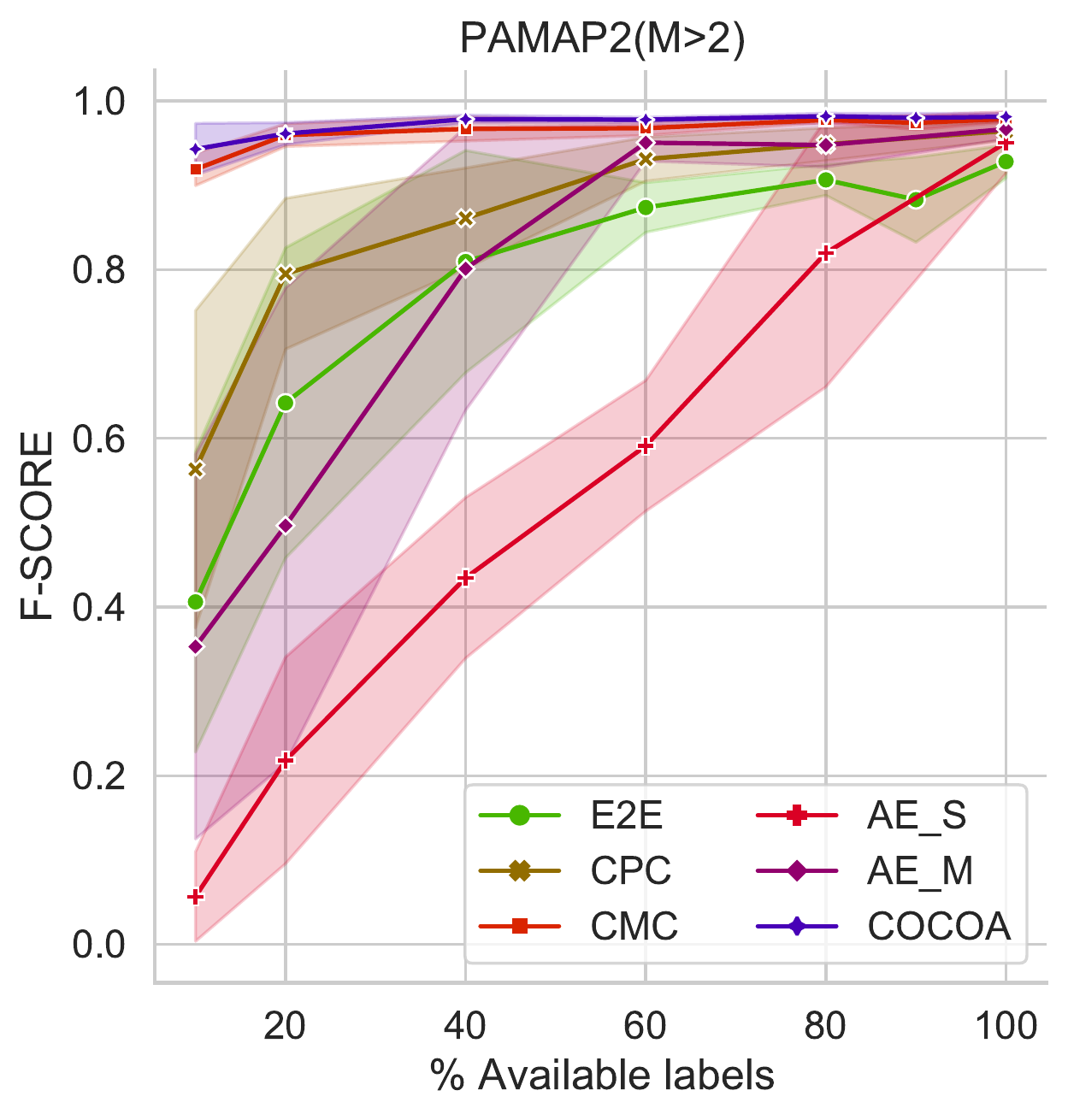}\label{fig:pamap_mlb}}
    \subfigure[]{\includegraphics[width=.24\linewidth]{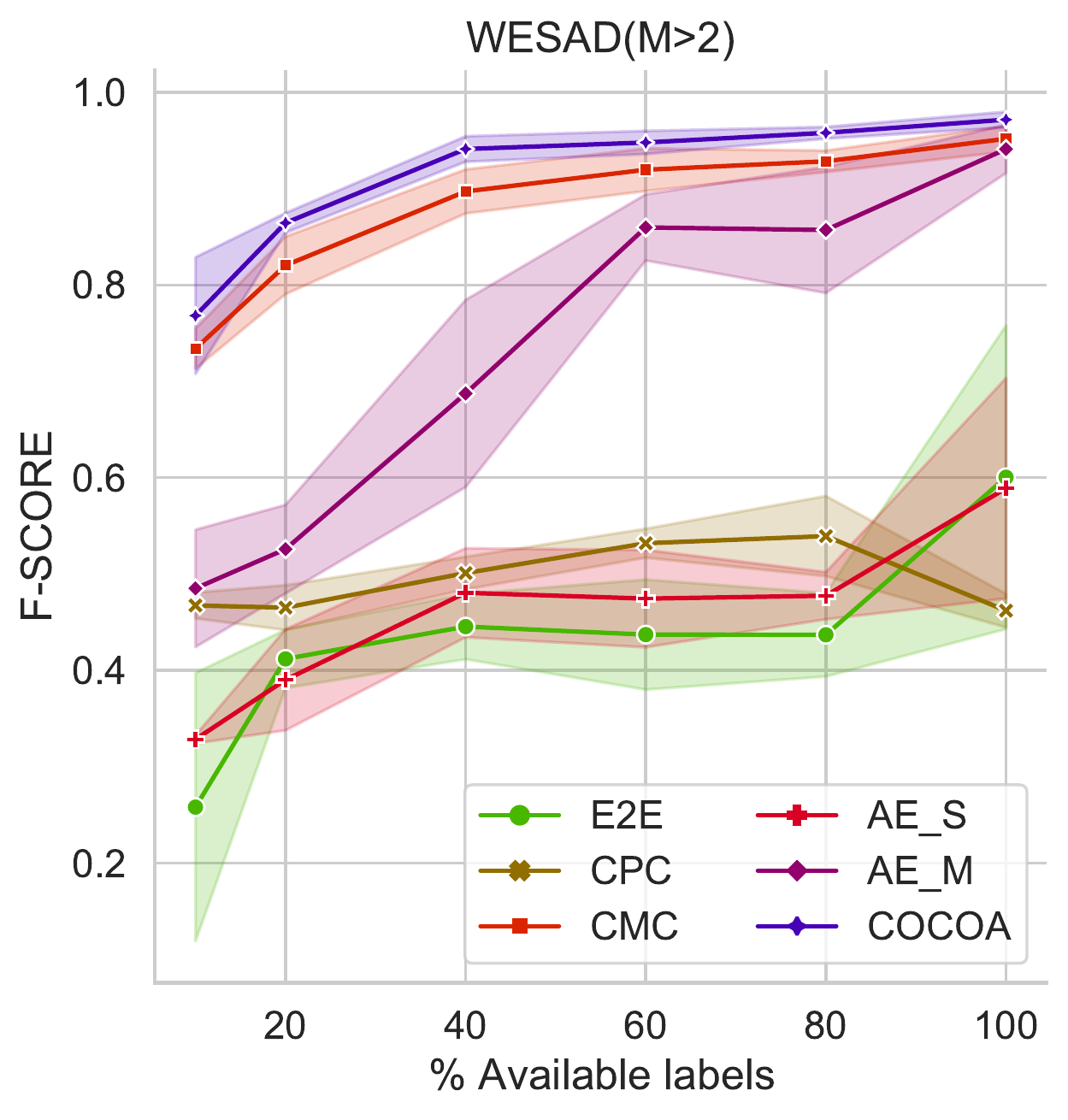}\label{fig:wesad_mlb}}
     \subfigure[]{\includegraphics[width=.24\linewidth]{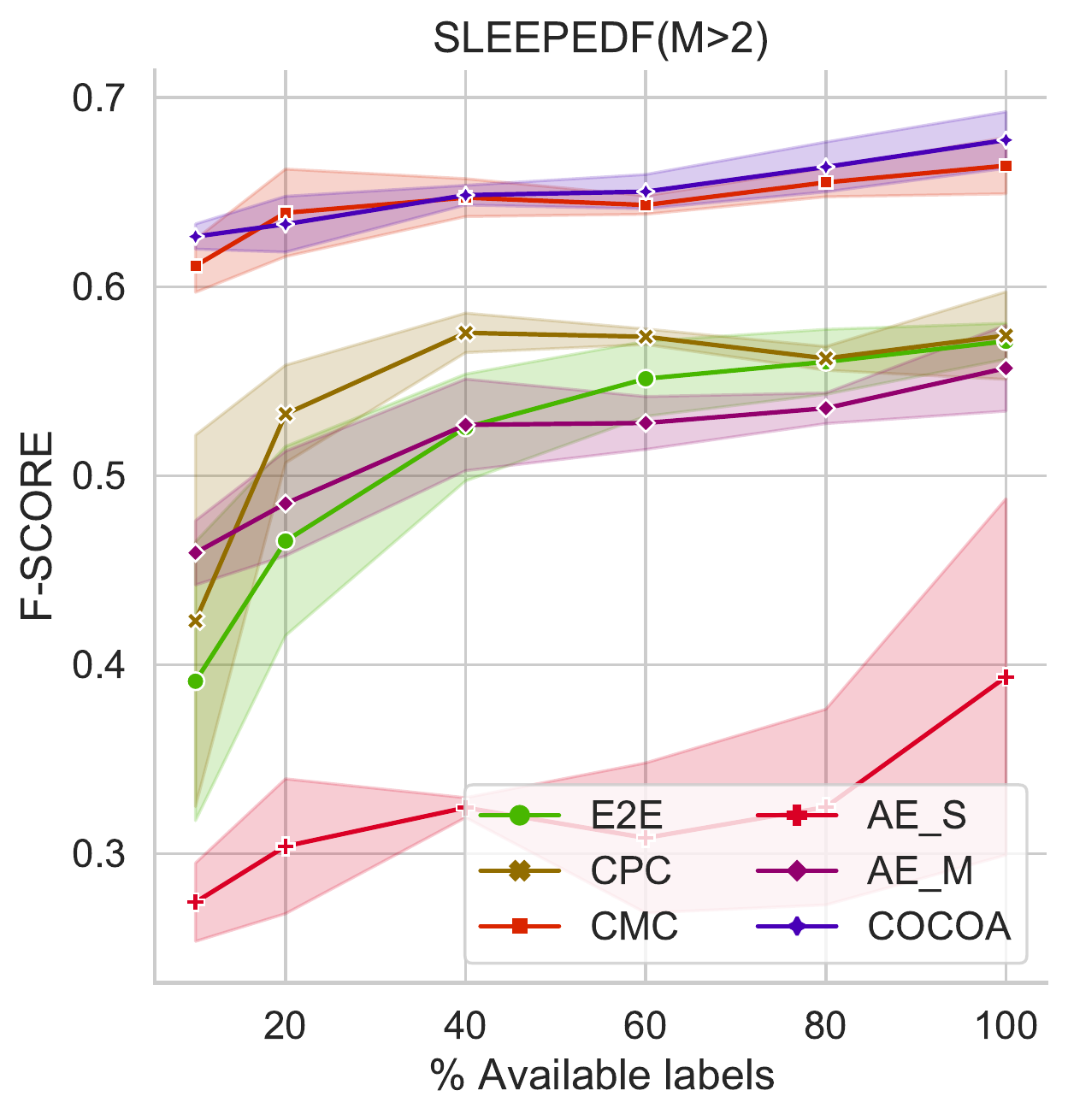}\label{fig:sleep_mlb}}
    \subfigure[]{\includegraphics[width=.24\linewidth]{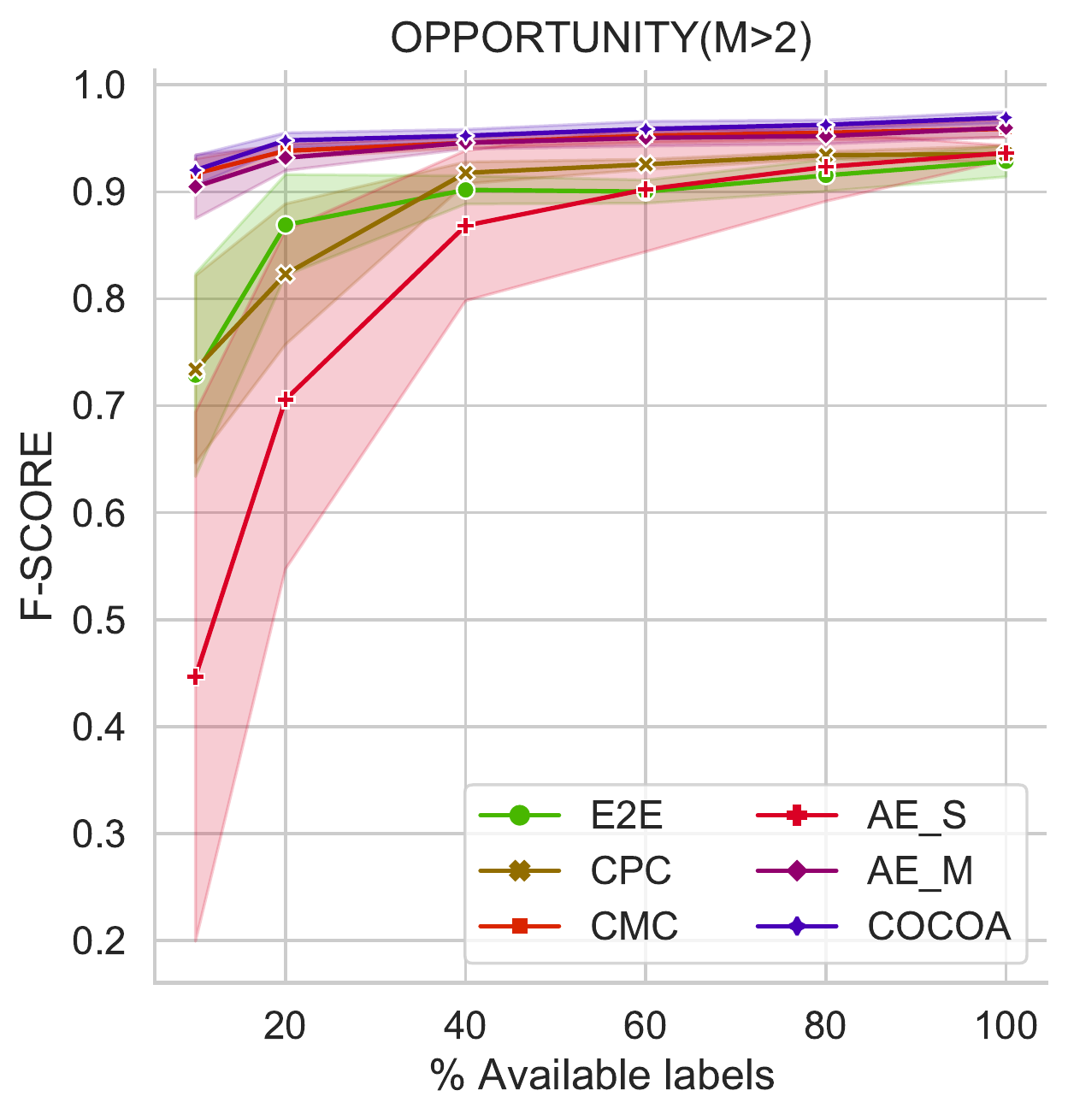}\label{fig:opp_mlb}}
    \caption{Effectiveness of \METHOD\ in low-data regimes where a small set of labeled data is available considering two (a,b,c,d) and more than two modalities (e,f,g,h). 
    }
    \label{fig:label_all}
\end{figure*}

\subsection{Data Efficiency}\label{Lab_effic}

We further studied the ability of \METHOD\ in learning general quality representations that can improve downstream tasks
in situations where only a small amount of labelled data is available. We conducted extensive experiments where the size of labelled data varies between $10$\% to $100$\% of the available labels for each dataset. 
For this experiment, we trained the classifier over self-supervised encoders by using only a subset of labelled data. The encoders are pre-trained without any labels and only fine-tuned along with the classifier based on the subset of labels. 
Based on Figures \ref{fig:pamap_lb}-\ref{fig:opp_lb}, \METHOD\ can reach the upper bound of the end-to-end supervised baseline by training upon only $10$\% of labelled data for the PAMAP2, WESAD and SLEEPEDF datasets, and using less than only $20$\% of the labels for the Opportunity dataset. The higher standard deviation in SLEEPEDF dataset \ref{fig:sleep_lb} is because of the various capability of selected sensors in predicting downstream task. 

Moreover, in case of multiple sensors data ($M>2$), Figures \ref{fig:pamap_mlb}-\ref{fig:opp_mlb}) confirm the effectiveness and efficiency of \METHOD\ compared to the fully-supervised ($E2E$) and the other cross-modal baselines (ie. $CMC$, $CPC$, and both autoencoders) models. \METHOD\ can reach an F-score of 96\%, 77\%, 63\% and 92\% in PAMAP2, WESAD, SLEEPEDF and Opportunity datasets, respectively while consuming only 10\% of labelled data. This shows about 10\%, 25\%, 14\% and 11\% improvement across datasets, respectively, compared to when only two modalities take part in contrastive learning. As it is shown, $AE-S$ achieve the highest performance of fully supervised baseline only if they are given the full labelled data in PAMAP2, WESAD and Opportunity datasets. Moreover, depending on the dataset, $AE-M$ can achieve more than the best supervised outcome upon the availability of at least $60$\%, $40$\%, $40$\% and $20$\% of the labeled dataset. Similarly, $CPC$ can achieve the $E2E$ supervised's prerformance by consuming $60$\% of labels in PAMAP2 and Opportunity datasets and $20$\% of annotations in WESAD dataset.


\begin{figure*}[]
\centering
     \subfigure[][]{
    \label{fig:two_pamap}
     \includegraphics[width=0.24\linewidth]{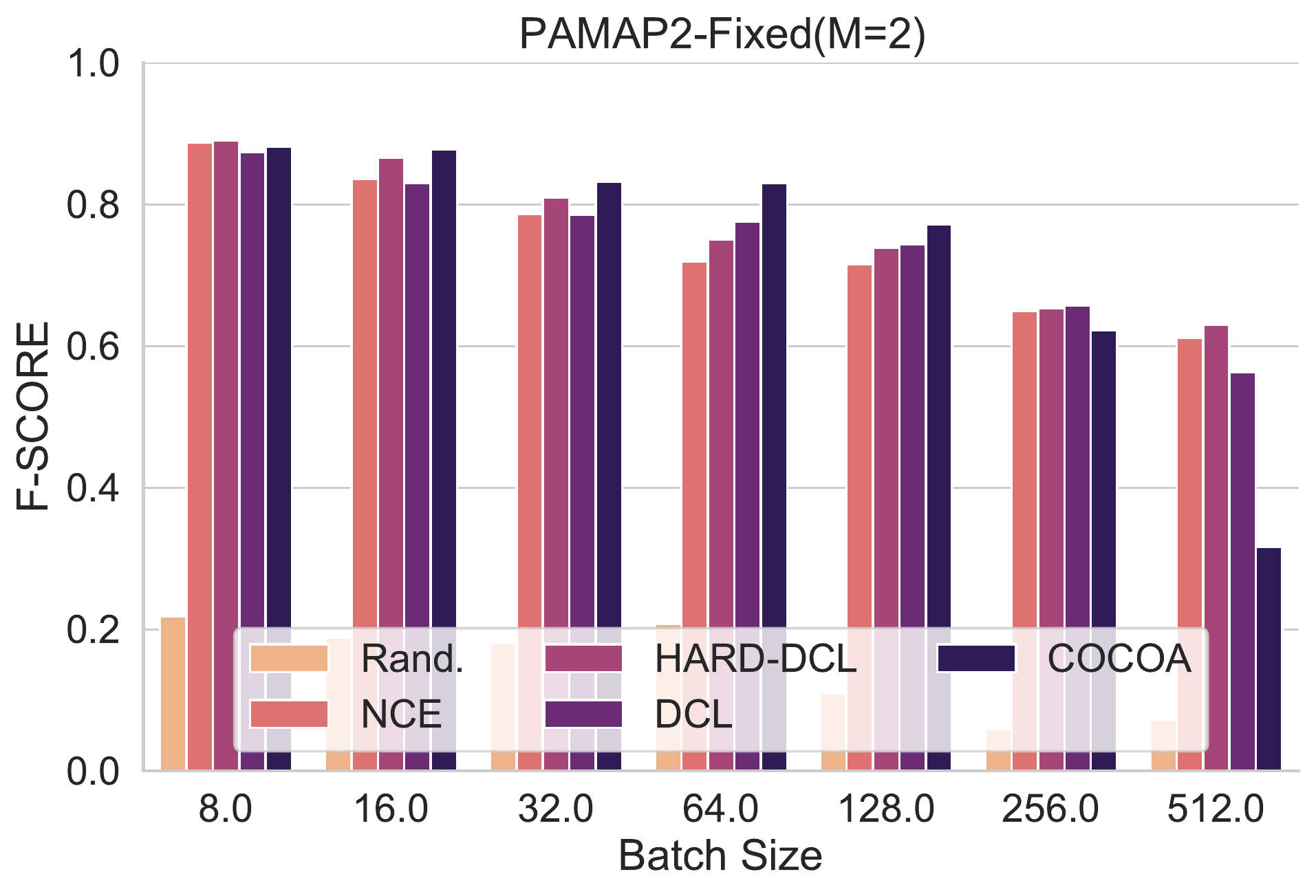}}
     \subfigure[][]{
    \label{fig:two_wesad}
     \includegraphics[width=0.24\linewidth]{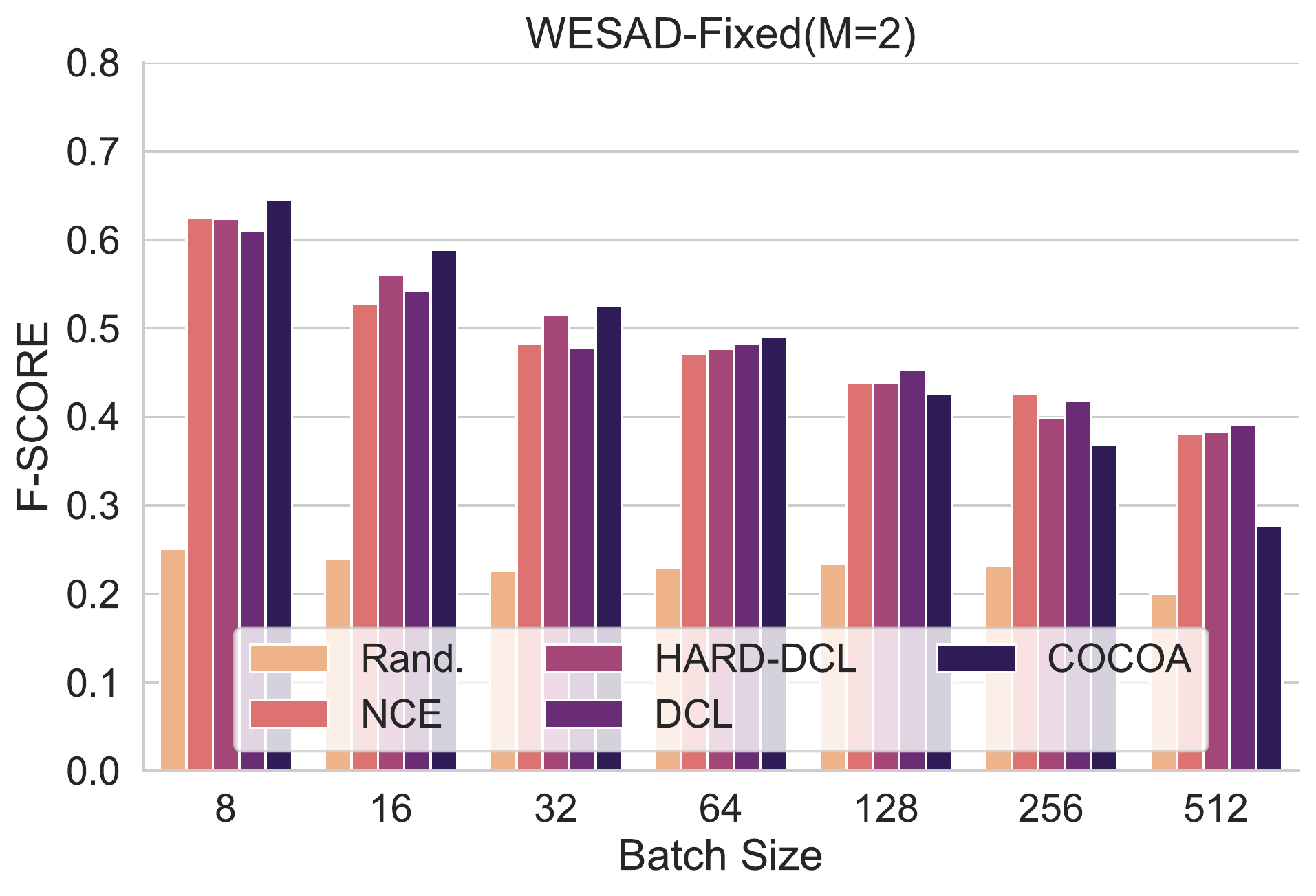}}
     \subfigure[][]{
     \label{fig:two_sleep}
     \includegraphics[width=0.24\linewidth]{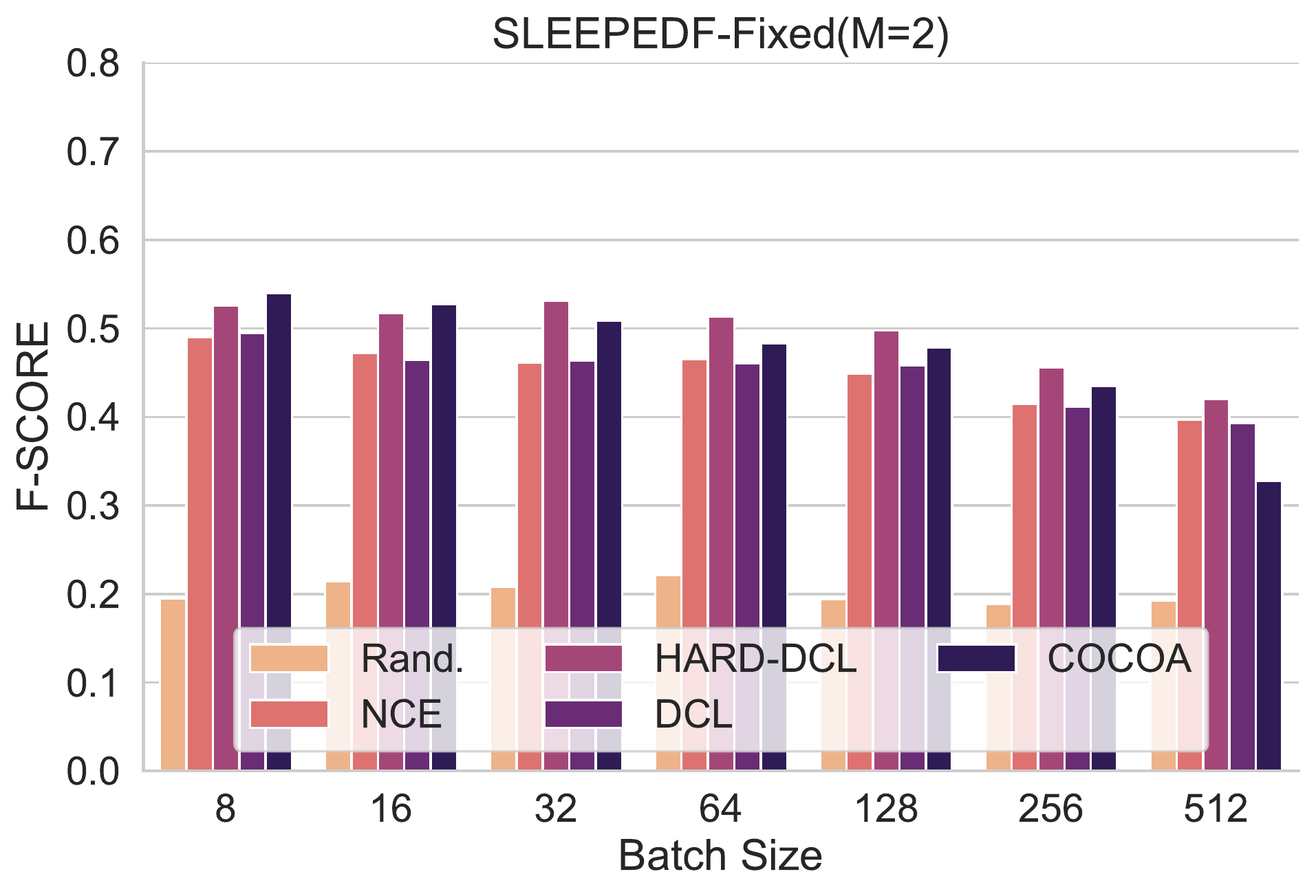}}
     \subfigure[][]{
     \label{fig:two_opp}
     \includegraphics[width=0.24\linewidth]{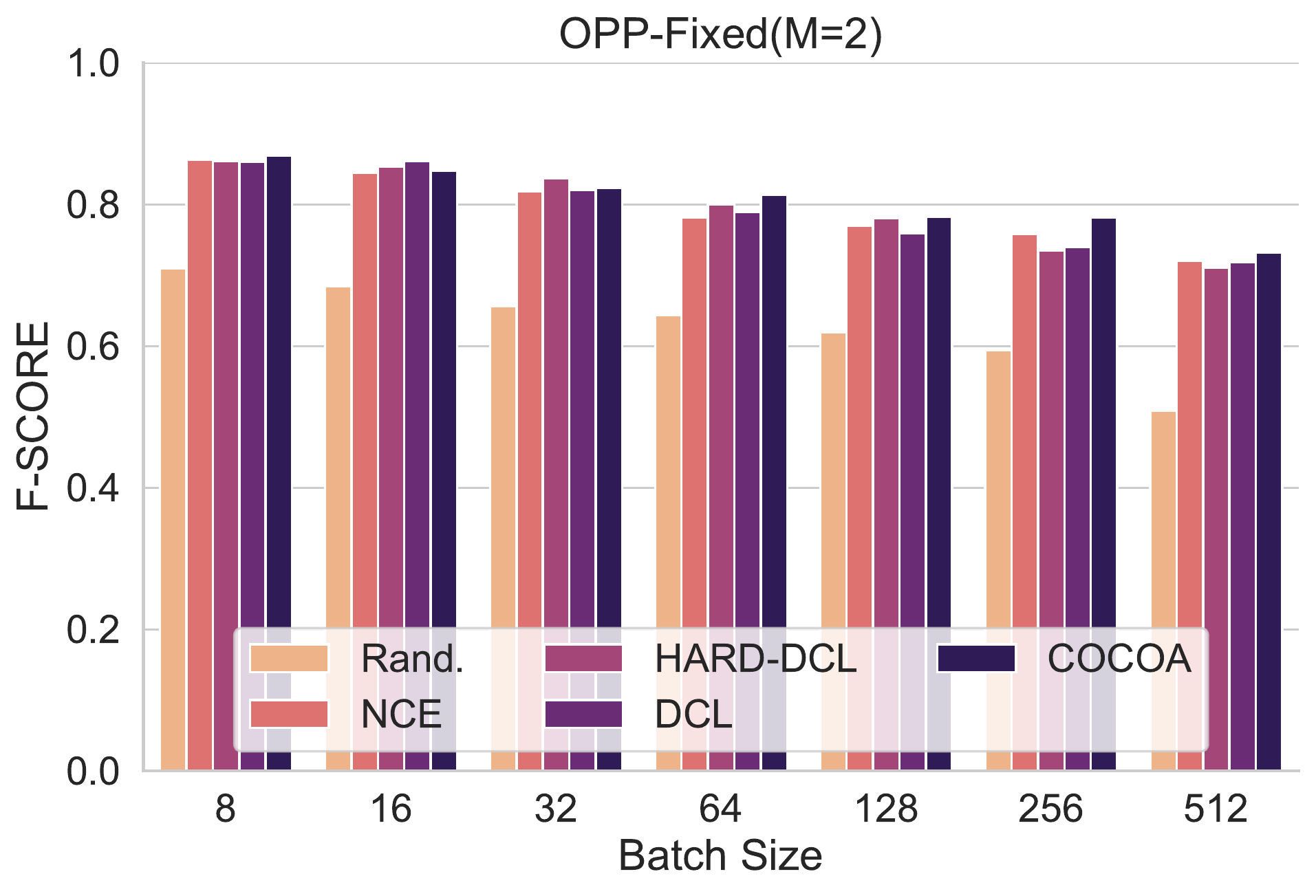}}
     \\
    \subfigure[][]{
     \label{fig:full_pamap}
     \includegraphics[width=0.24\linewidth]{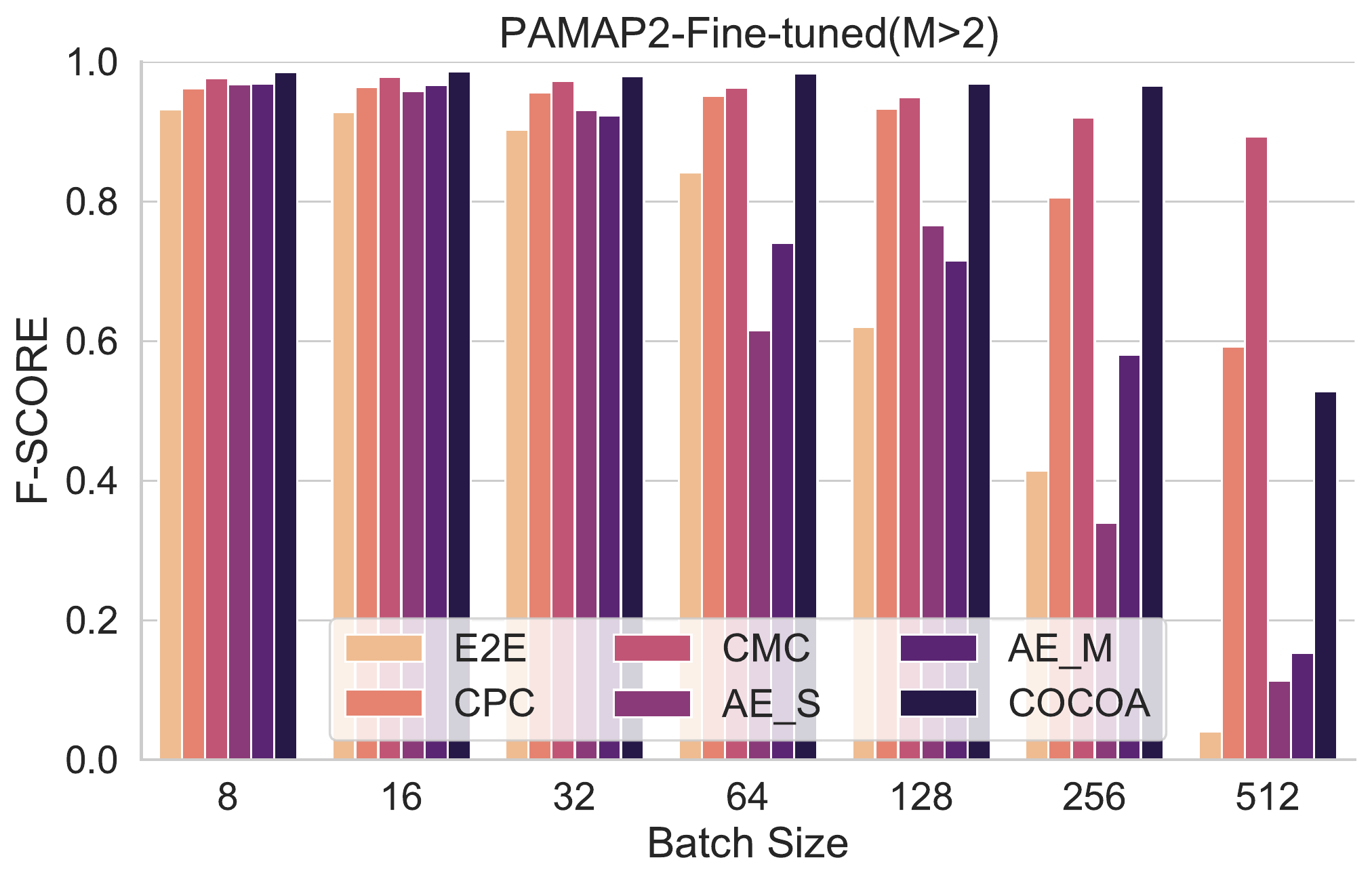}}
    \subfigure[][]{
    \label{fig:full_wesad}
     \includegraphics[width=0.24\linewidth]{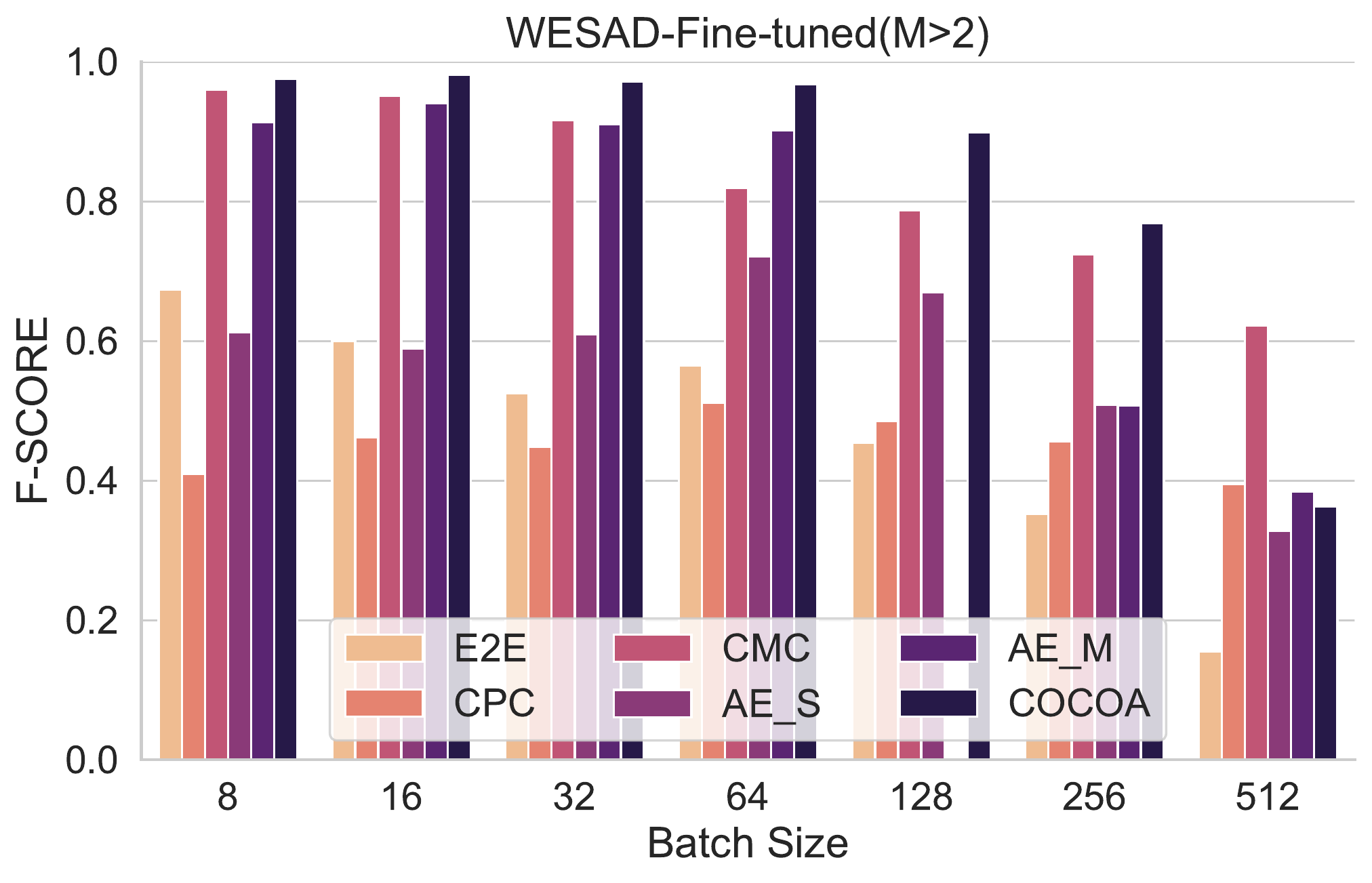}}
     \subfigure[][]{
    \label{fig:full_sleep}
     \includegraphics[width=0.24\linewidth]{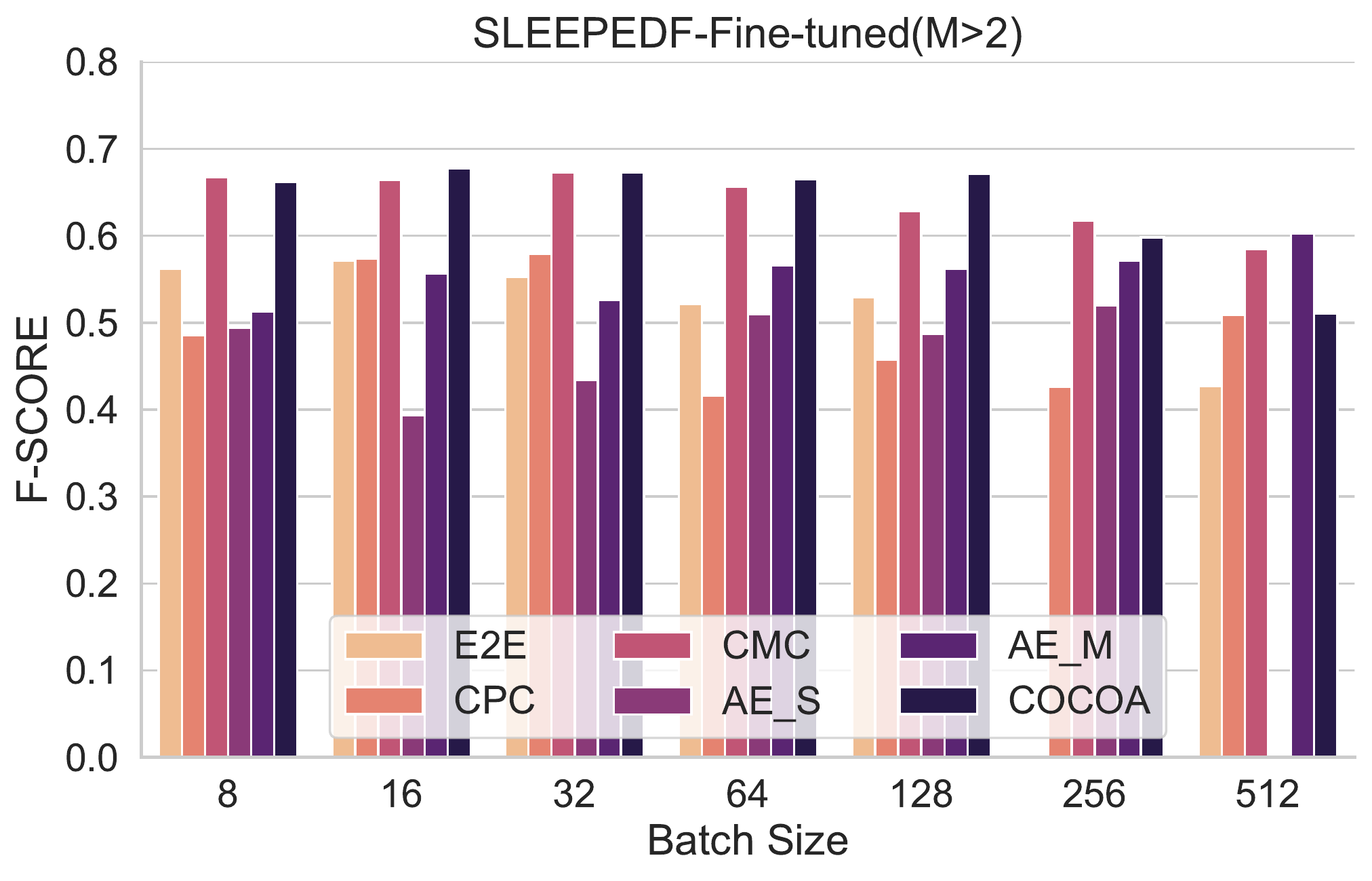}}
     \subfigure[][]{
     \label{fig:full_opp}
     \includegraphics[width=0.24\linewidth]{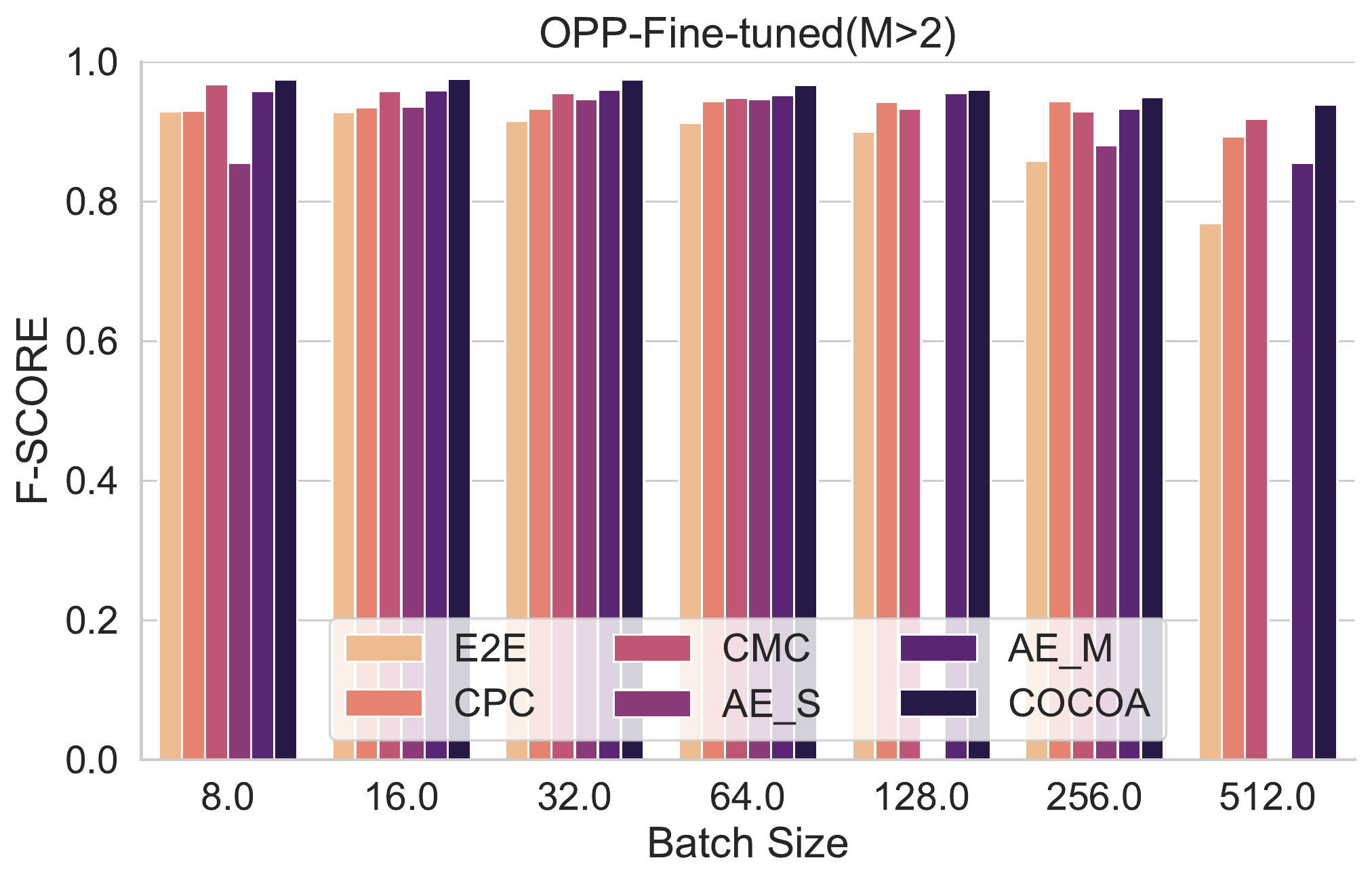}}
     
    \caption{Sensitivity analysis. \METHOD\ is compared against other baselines with fine-tuned encoders with dual- (figures a,b,c,d) and multiple modalities (figures e,f,g,h) and over range of batch sizes across PAMAP2, WESAD, SLEEPEDF, and Opportunity datasets.}
    \label{fig:all_baselines}
\end{figure*}

\subsection{Batch Size Sensitivity Analysis}
\label{sec:sens_anly}

Figure \ref{fig:all_baselines} provides comparison across a range of batch sizes with Fixed (Figures \ref{fig:two_pamap}-\ref{fig:two_opp}) and Fine-tuned (Figures\ref{fig:full_pamap}-\ref{fig:full_opp}) Encoders. The Figures demonstrate that learnt representations provide a significant contribution toward improving generalization on the downstream task compared to the supervised lower bound, fully supervised classifier with fixed randomly initiated encoder (Fixed), and the end-to-end fully supervised approach (E2E). In the case of contrasting only two modalities, \METHOD\ offers a higher F-score for the majority of the batch sizes except for larger batch sizes (i.e. 512) across PAMAP2 and WESAD datasets. For the SLEEPEDF and Opportunity datasets, it is highly competitive with the \textit{DCL, Hard-DCL}. As we explained earlier, $DCL$ and $Hard-DCL$ are the only methods that proposed an approach to a debiased objective function to reduce the effect of fake negative samples. Similarly, in case of more than two modalities, \METHOD\ outperform the other baselines across most batch sizes. \METHOD\ and $CMC$ as the only cross-modal contrastive approaches provide larger improvements compared to other baselines.
Furthermore, in both dual and multimodal modes, the sensitivity analysis shows all methods offer superior performance for smaller batch sizes and the F-score degrades as the batch size increase. 

\begin{figure*}[]
\centering
     \subfigure[][]{
    \label{fig:ssl_pamap}
     \includegraphics[width=0.24\linewidth]{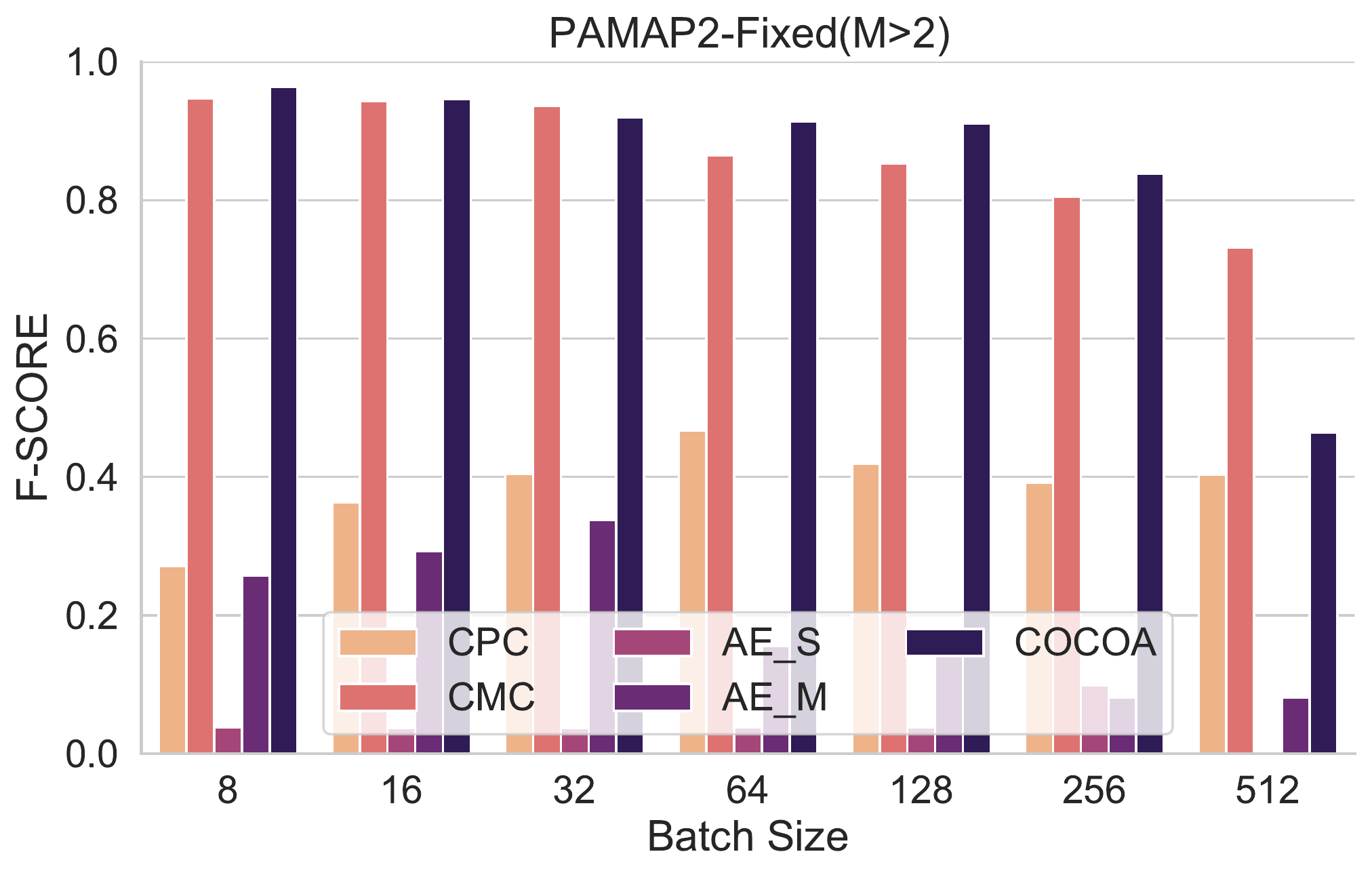}}
     \subfigure[][]{
    \label{fig:ssl_wesad}
     \includegraphics[width=0.24\linewidth]{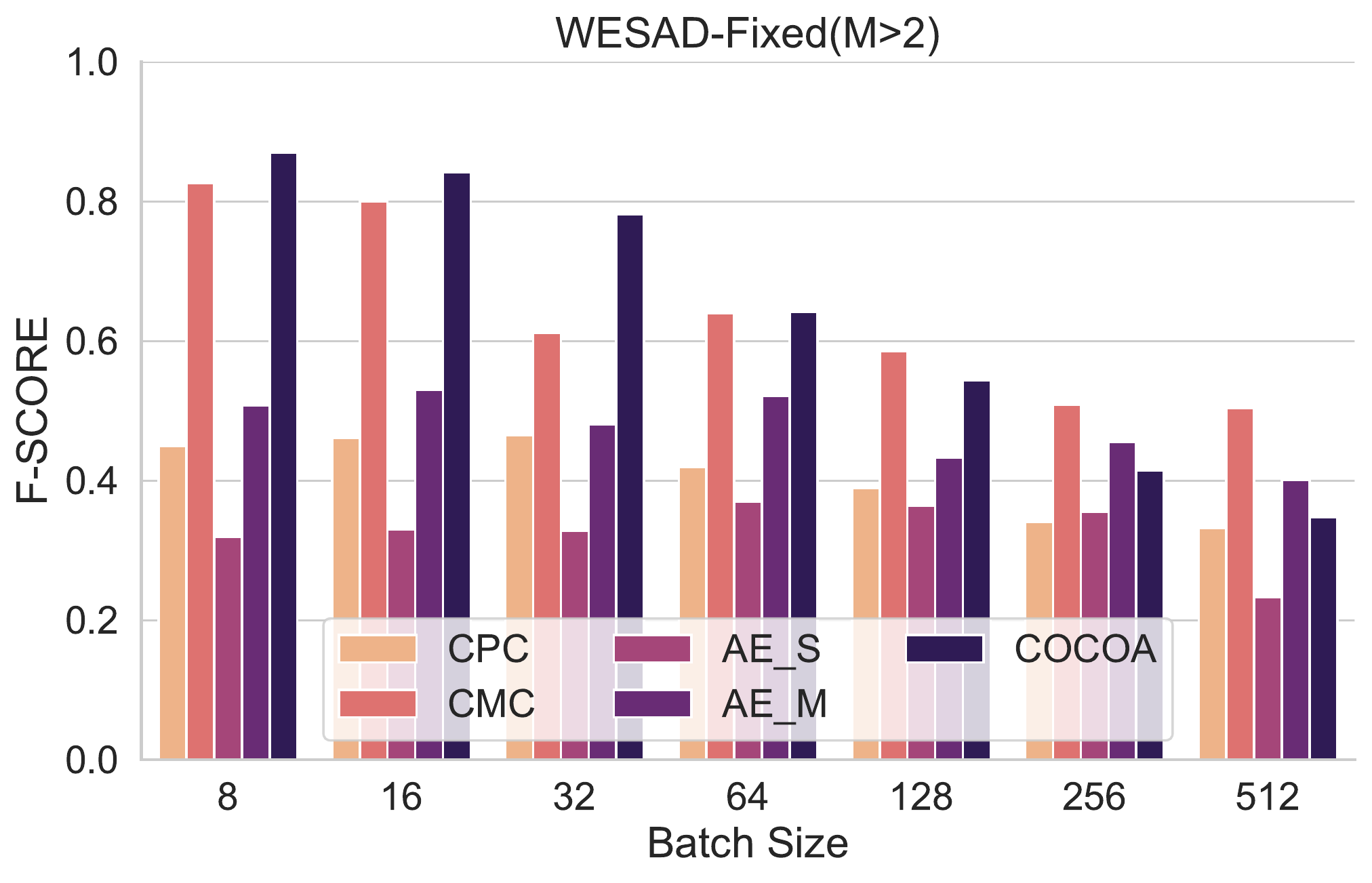}}
     \subfigure[][]{
     \label{fig:ssl_sleep}
     \includegraphics[width=0.24\linewidth]{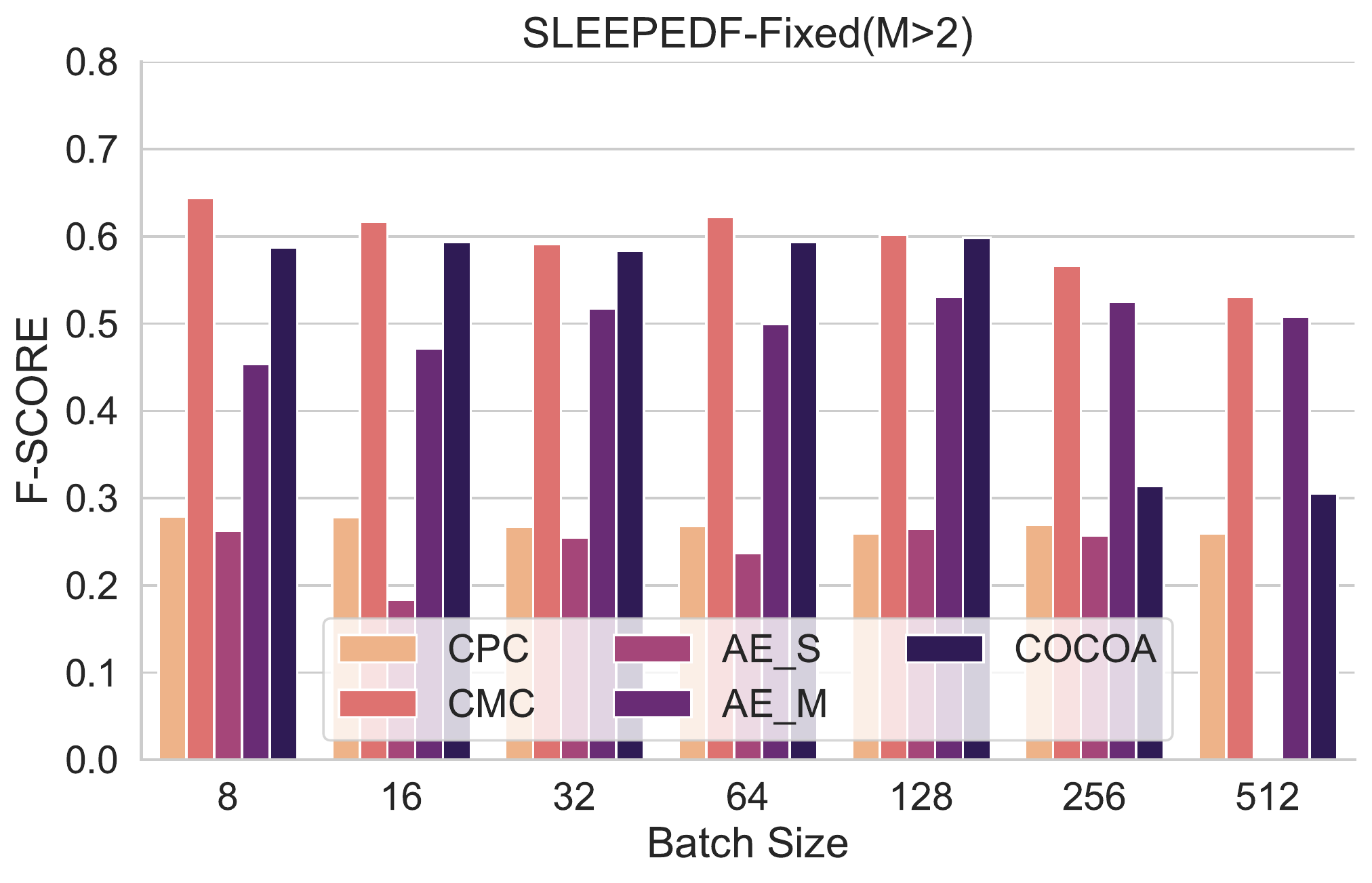}}
     \subfigure[][]{
     \label{fig:ssl_opp}
     \includegraphics[width=0.24\linewidth]{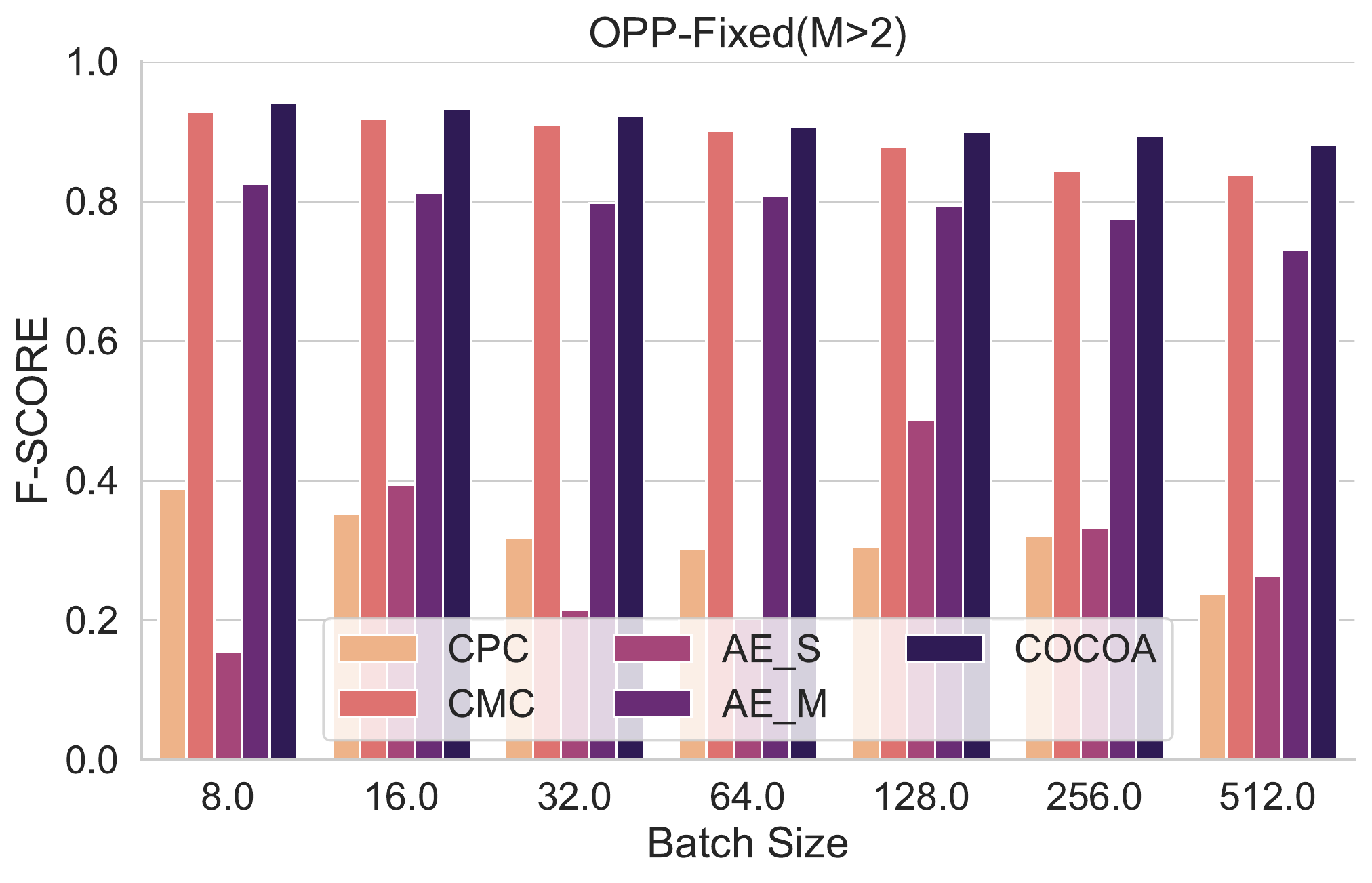}}
     \caption{Comparing the effect of batch size with fixed encoders across (a) PAMAP2, (b) WESAD, (c) SLEEPEDF and (d) Opportunity dataset.}
     \label{fig:fixed_sensitivity}
\end{figure*}

Further evaluation results are provided in Figure \ref{fig:fixed_sensitivity} to compare the effectiveness of each method with fixed encoders across datasets with more than two modal data. In this scenario, the pre-trained encoders are frozen and are not fine-tuned during the classification. This set of comparisons is necessary to study the effectiveness of the self-supervised methods as they only measure the performance of the pure self-supervised part by classifying the representation learnt by each encoder. 
According to Figures \ref{fig:ssl_pamap}, \ref{fig:ssl_wesad}, and \ref{fig:ssl_sleep}, representations learnt using \METHOD\ are more distinguished and provide quality information to the classifier. 
As the figures show, \METHOD-based encoders can extract more quality representations compared to the other baselines. 

As we expect, the supervised baseline, auto-encoders and $CPC$ are not dependent on the batch size since the training goal does not involve any comparison between samples within the batch. However, the performance of $CMC$ and \METHOD\ decrease with larger batch sizes. this confirms the compatibility of sensor data contrastive models with smaller batch sizes \cite{deldari2021tscp2}. 
 In computer vision, however, larger batch sizes commonly lead to higher accuracy by providing a more diverse set of contrasting opportunities \cite{chen2020simple}. In time-series data considered here (human activity recognition, sleep stage detection, and emotion recognition applications), however, larger batch sizes do not happen to be as beneficial. We hypothesise this could be due to the less number of classes defined in these applications compared with natural language or computer vision domains. Furthermore, in applications such as behavioural or human activity recognition, classes are likely to \textit{repeat} over time. Consequently, larger batch sizes are more likely to include more false-negative samples that can degrade the performance of contrastive objective function as it will mistakenly separate the semantically related instances within the shared representation space.

\subsection{Visualising the Distribution of Latent Space}
\label{sec:rep_vis}
\begin{figure}[ht]
\subfigure[][]{
     \label{fig:uci_tsne_raw}
     \includegraphics[width=0.20\linewidth]{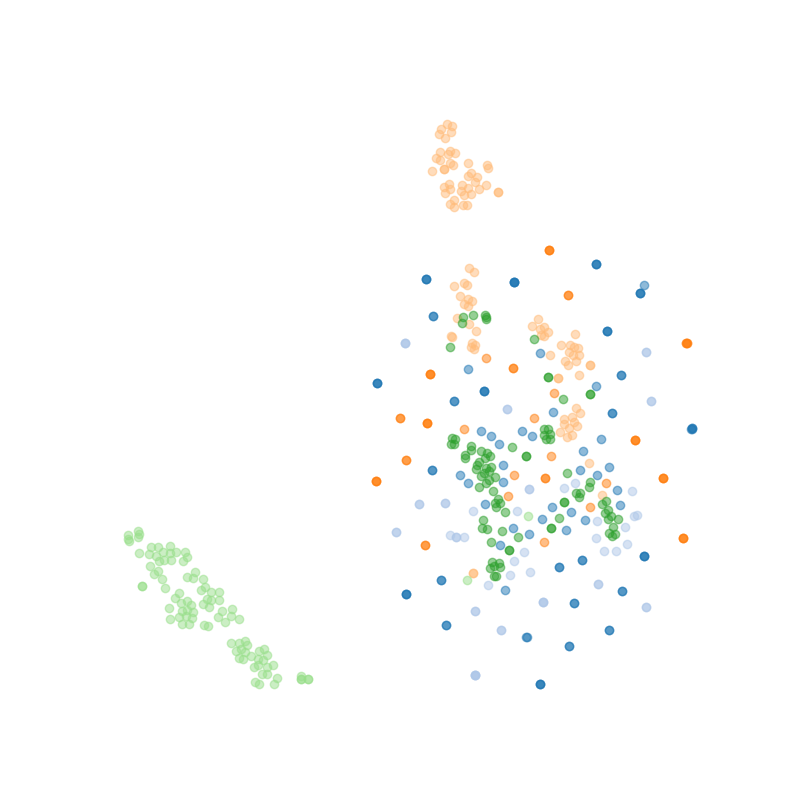}}
\subfigure[][]{
     \label{fig:uci_tsne_ssl}
     \includegraphics[width=0.20\linewidth]{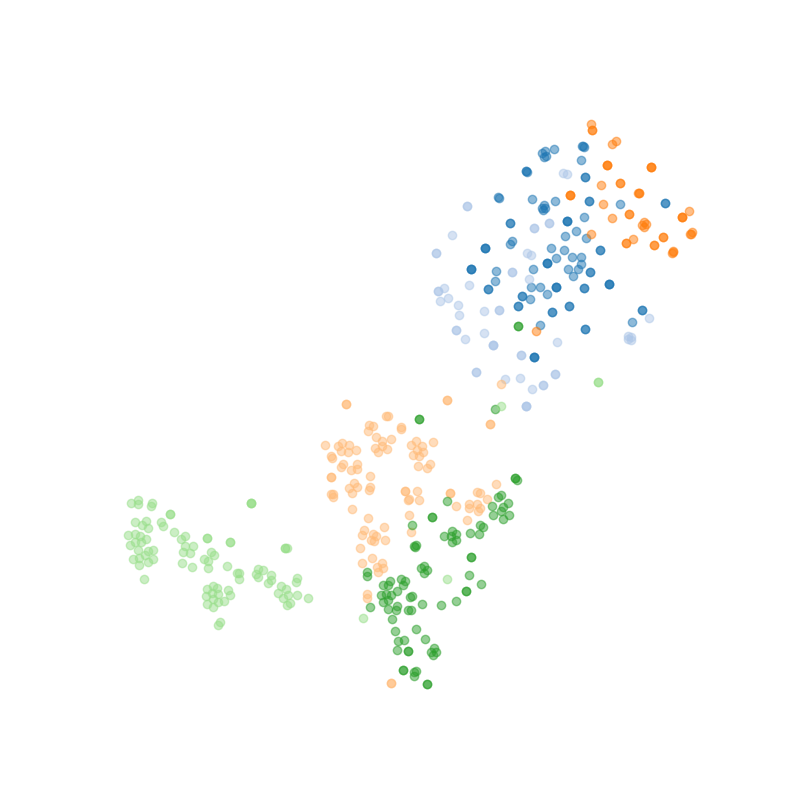}}
\subfigure[][]{
     \label{fig:uci_tsne_fine}
     \includegraphics[width=0.20\linewidth]{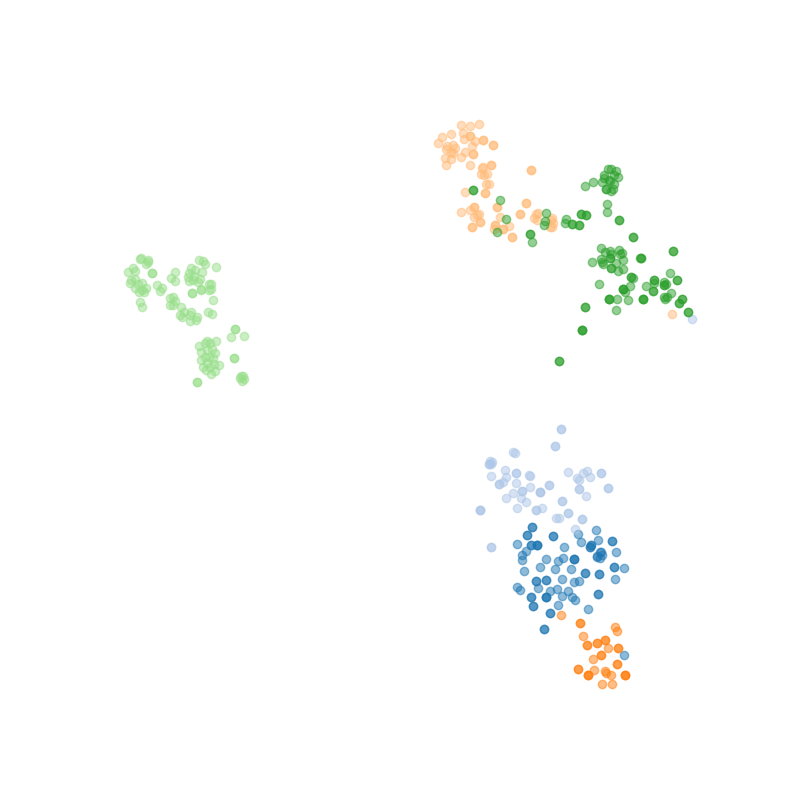}}
\subfigure[][]{
     \includegraphics[width=0.15\linewidth]{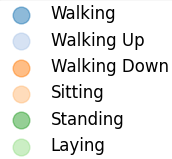}}\\
\caption{UCIHAR. t-SNE Visualisation of: (a) Original input data, embeddings from (b) SSL, and (c) Fine-tuned encoders. }
\label{fig:uci_tsne}
\end{figure}

\begin{figure}[ht]    
\subfigure[][]{
     \label{fig:pamap_tsne_raw}
     \includegraphics[width=0.2\linewidth]{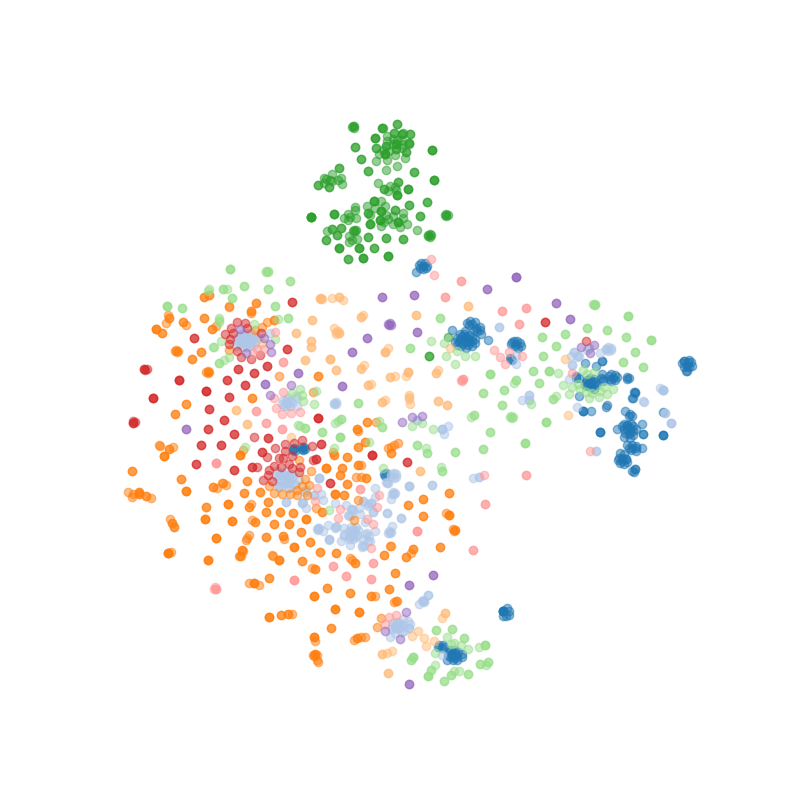}}
\subfigure[][]{
     \label{fig:pamap_tsne_ssl}
     \includegraphics[width=0.2\linewidth]{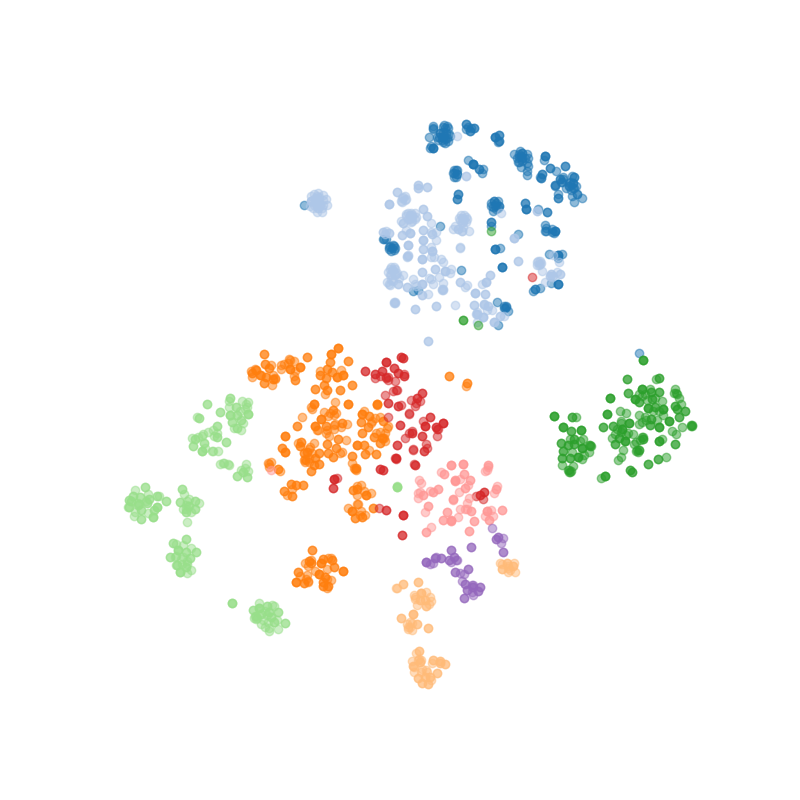}}
\subfigure[][]{
     \label{fig:pamap_tsne_fine}
     \includegraphics[width=0.2\linewidth]{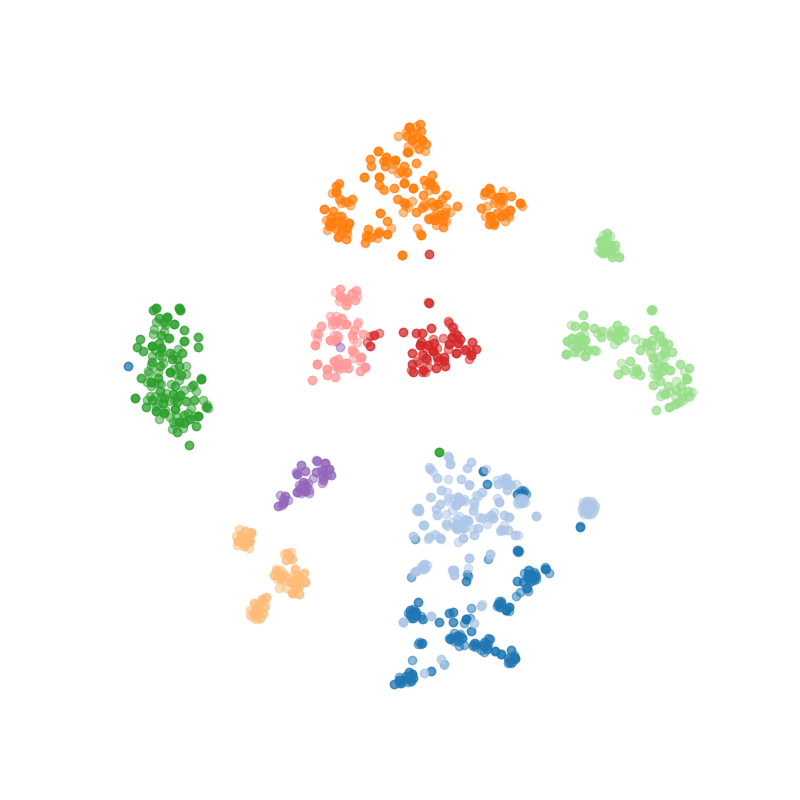}}
     \subfigure[][]{
     \includegraphics[width=0.15\linewidth]{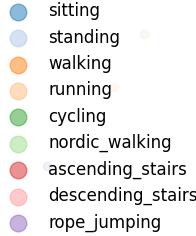}}\\
 \caption{PAMAP2 Dataset. t-SNE Visualisation of: (a) Original input data, embeddings from (b) SSL, and (c) Fine-tuned encoder.}
\label{fig:pamap_tsne}
\end{figure}

\begin{figure}[ht]    
\subfigure[][]{
     \label{fig:sleep_tsne_raw}
     \includegraphics[width=0.2\linewidth]{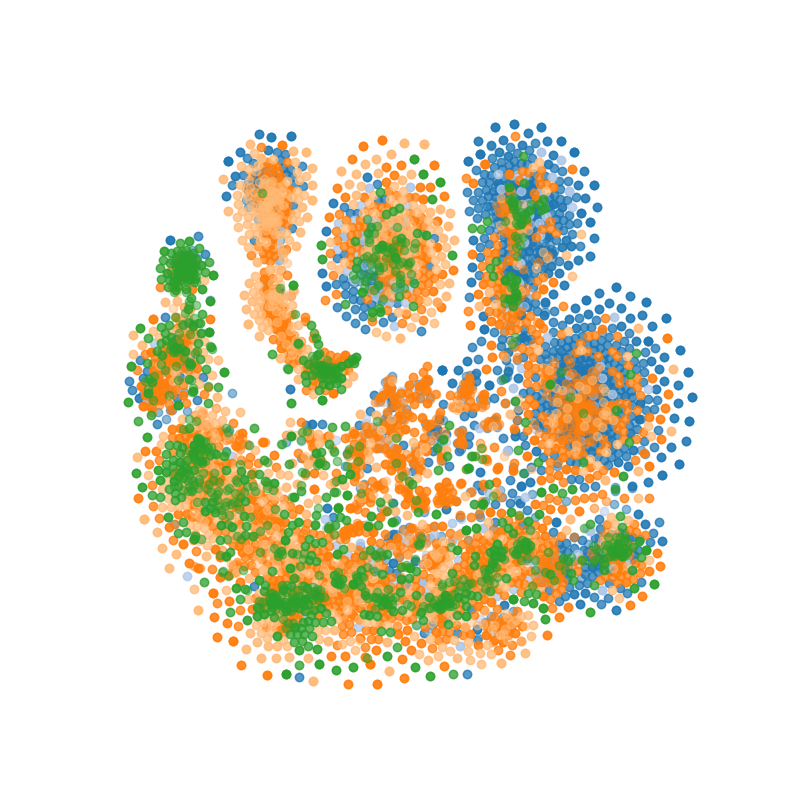}}
\subfigure[][]{
     \label{fig:sleep_tsne_ssl}
     \includegraphics[width=0.2\linewidth]{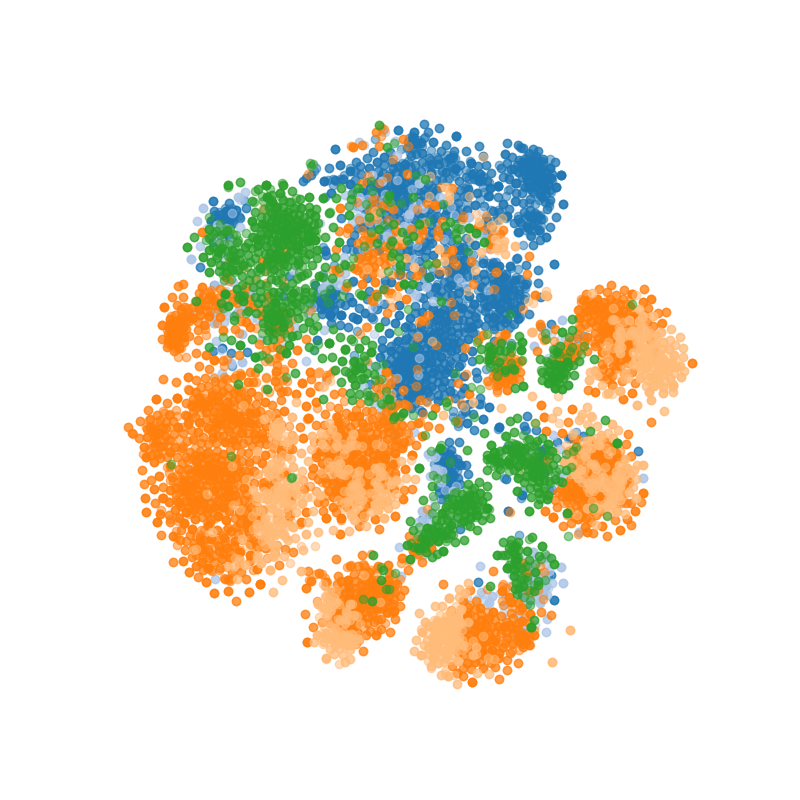}}
\subfigure[][]{
     \label{fig:sleep_tsne_fine}
     \includegraphics[width=0.2\linewidth]{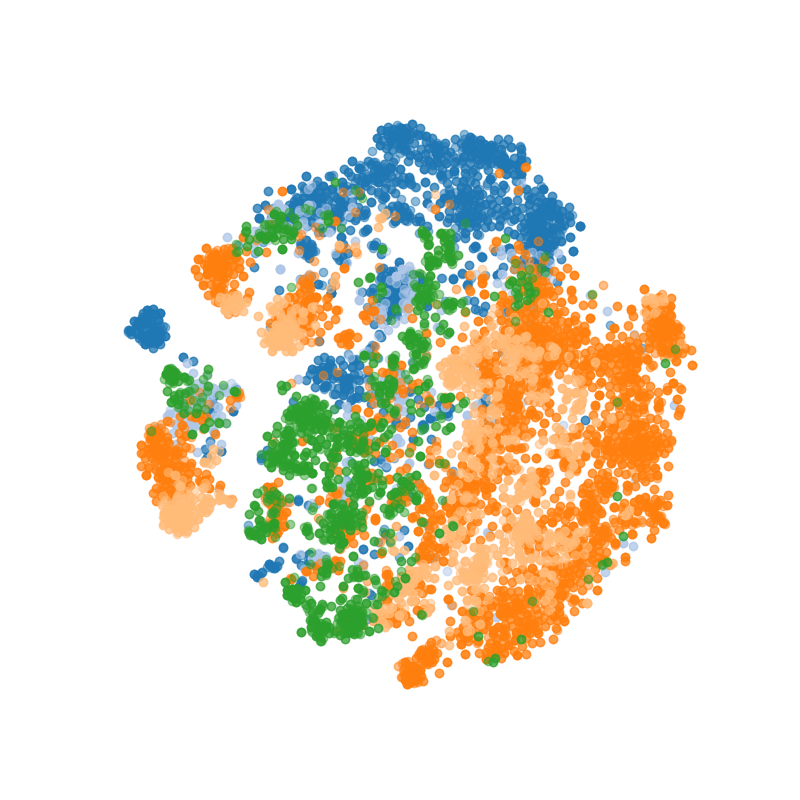}}
     \subfigure[][]{
     \includegraphics[width=0.1\linewidth]{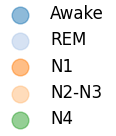}}\\
\caption{SLEEPEDF Dataset. t-SNE Visualisation of: (a) Original input data, embeddings from (b) SSL, (c) Fine-tuned encoder.}
\label{fig:sleep_tsne}
\end{figure}

\begin{figure}[ht]
\subfigure[][]{
     \label{fig:wesad_tsne_raw}
     \includegraphics[width=0.2\linewidth]{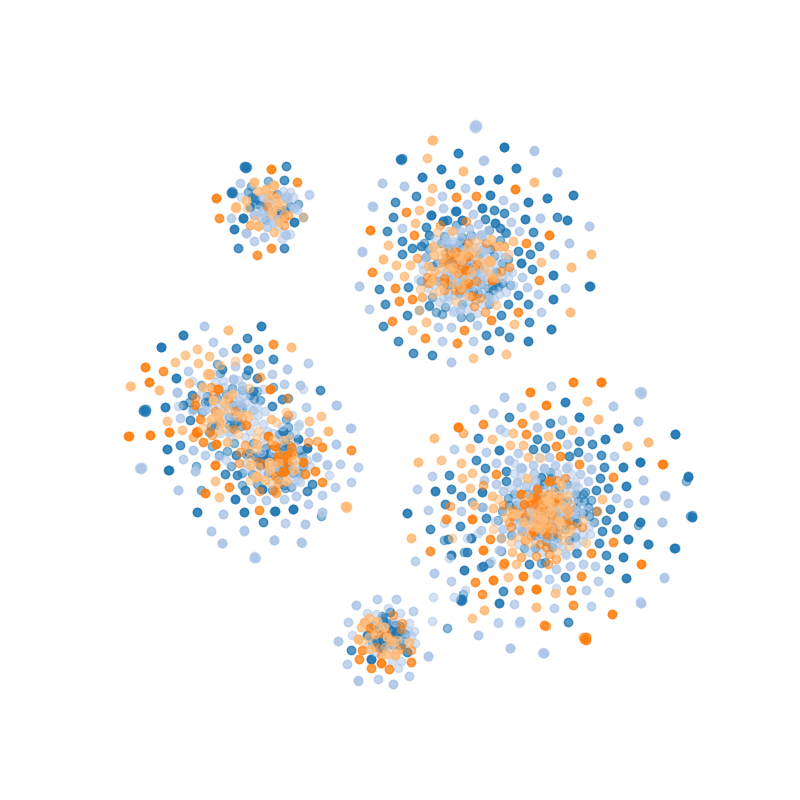}}
\subfigure[][]{
     \label{fig:wesad_tsne_ssl}
     \includegraphics[width=0.2\linewidth]{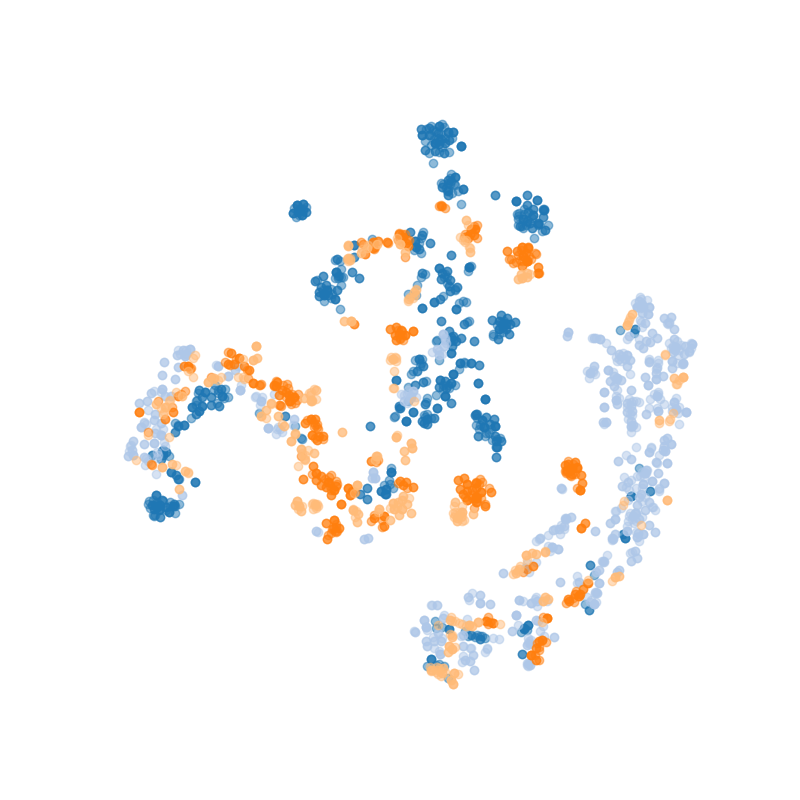}}
\subfigure[][]{
     \label{fig:wesad_tsne_fine}
     \includegraphics[width=0.2\linewidth]{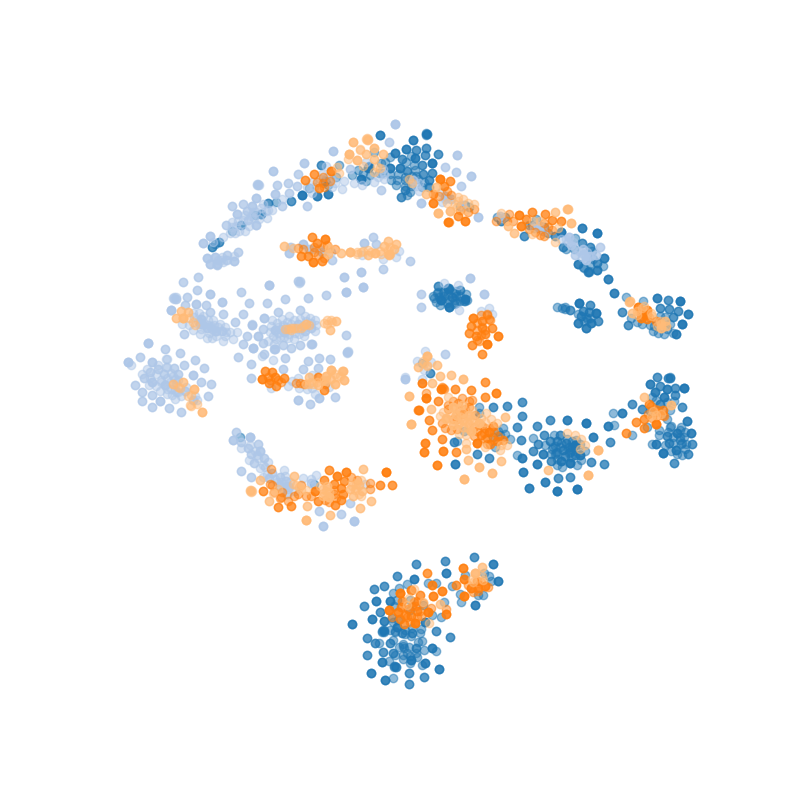}}
     \subfigure[][]{
     \includegraphics[width=0.13\linewidth]{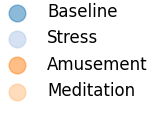}}\\
\caption{WESAD Dataset. t-SNE Visualisation of: (a) Original input data, embeddings from (b) SSL, and (c) Fine-tuned encoder.}
\label{fig:wesad_tsne}
\end{figure}

\begin{figure}[ht]
\subfigure[][]{
     \label{fig:opp_tsne_raw}
     \includegraphics[width=0.2\linewidth]{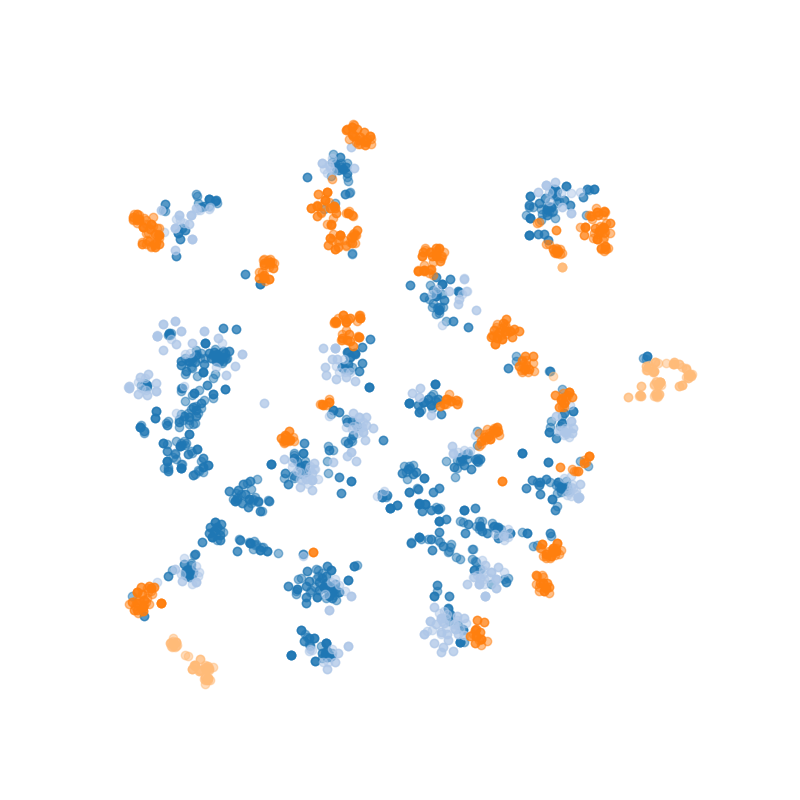}}
\subfigure[][]{
     \label{fig:opp_tsne_ssl}
     \includegraphics[width=0.2\linewidth]{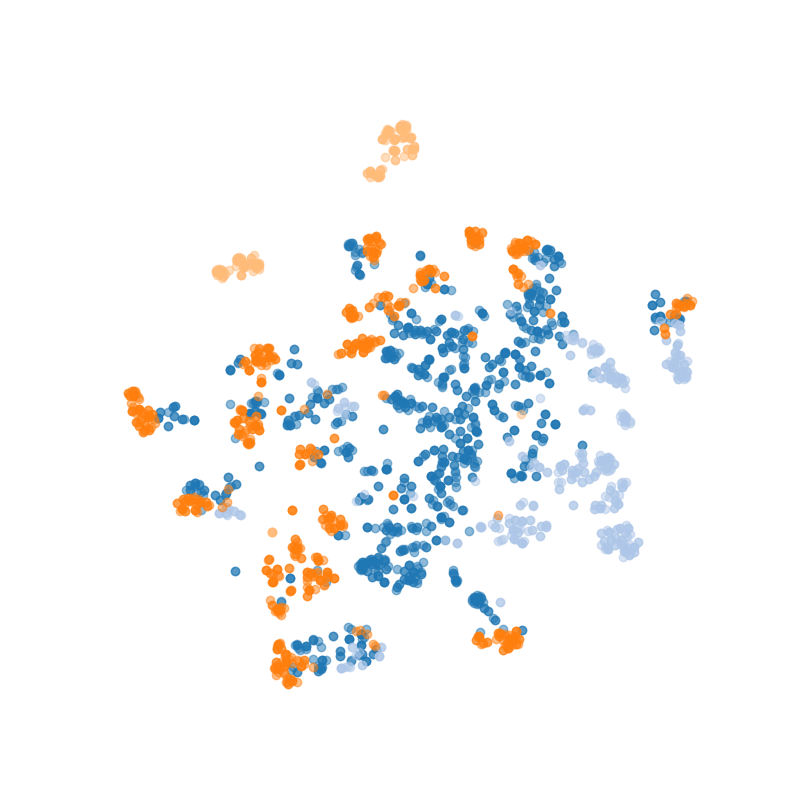}}
\subfigure[][]{
     \label{fig:opp_tsne_fine}
     \includegraphics[width=0.2\linewidth]{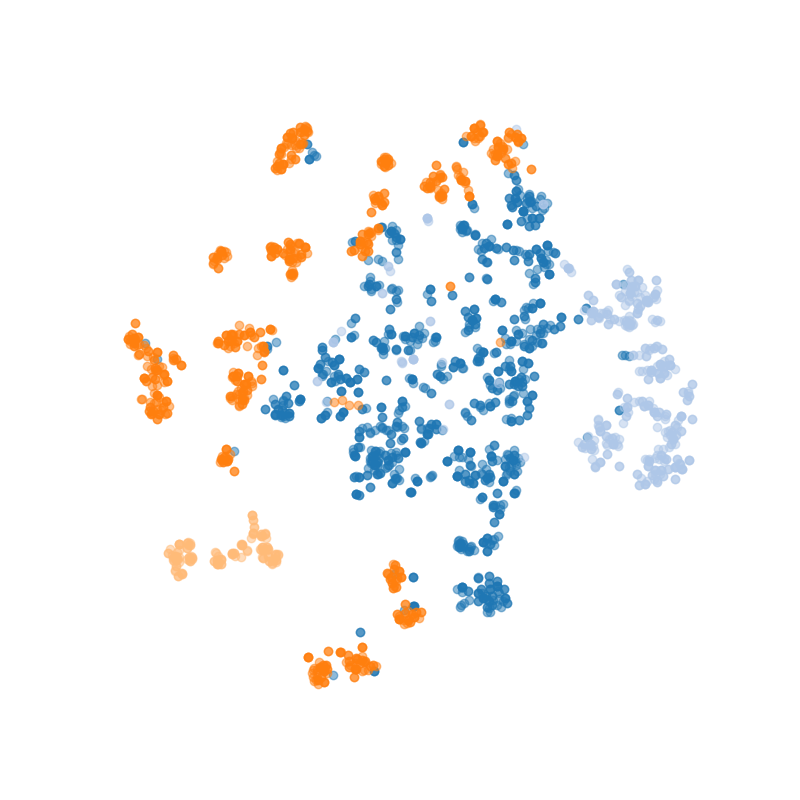}}
     \subfigure[][]{
     \includegraphics[width=0.13\linewidth]{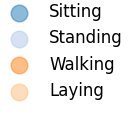}}\\
\caption{Opportunity Dataset. t-SNE Visualisation of: (a) Raw data, and embeddings from (b) SSL, and (c) Fine-tuned encoder.}
\label{fig:opp_tsne}
\end{figure}

As we discussed in previous sections, our experiments confirm that the representations learnt by \METHOD\ yield competitive and higher performance on classification tasks across studied datasets. To show the structure and distribution of raw data and corresponding embedding vectors, we visualise them in a 2-Dimensional space using t-SNE \cite{van2008tsnevisualizing}, a well-known dimensionality reduction technique \footnote{We employed SciKit-Learn implementation of t-SNE \cite{scikit-learn}.}. Figures \ref{fig:uci_tsne}-\ref{fig:opp_tsne} compare visualisation of 
\begin{itemize}
    \item (a): Raw input data,
    \item (b): SSL-representations which are the output of purely self-supervised encoders, and
    \item (c): Fine-tuned representations which are the output of fine-tuned pre-trained (SSL) encoders,
\end{itemize}
across UCIHAR, PAMAP2, SLEEPEDF, WESAD and Opportunity datasets, respectively. Each point in the plots indicates a $D$-dimensional time-series of length $W$, where $D$ equals the total number of channels (including all sensors) and $W$ represents the window size, as described in Section \ref{sec:datasets} for raw input plots (a), and represents the size of embedding vector in the SSL and fine-tuned plots (figures (b) and (c), respectively).

In UCIHAR dataset, representations of \textit{sitting, laying} and \textit{standing} poses are clearly distinguished from other activities using both self-supervised (Figure \ref{fig:uci_tsne_ssl}) and fine-tuned \ref{fig:uci_tsne_fine}) encoders. However, \textit{walking} and \textit{walking up/downstairs} are distinguished more clearly by the fine-tuned encoder.
Raw signals from PAMAP2 dataset are shown in Figure \ref{fig:pamap_tsne}. Apart from \textit{cycling} activity, other activities, specially \textit{walking} and \textit{nordic-walking} activities, are distributed almost all over the feature space. Although the distribution of PAMAP2's samples do not show well-defined clusters \ref{fig:pamap_tsne_raw}, both SSL (Figure \ref{fig:pamap_tsne_ssl}) and fine-tuned \METHOD\ (Figure \ref{fig:pamap_tsne_fine}) can successfully differentiate between all activities.

Upon inspection of the distribution of sample of SLEEPEDF dataset capturing different sleep stages, boundaries of emerged clusters are almost clear in the SSL encoder in Figure \ref{fig:sleep_tsne_ssl}, but more distinguishable in Fine-tuned \METHOD\, Figure \ref{fig:sleep_tsne_fine}. The distribution of learnt representations are almost aligned with previous observations in \cite{banville2021uncovering} as they are sequentially arranged considering their sleep stage labels: moving from awake state to Rapid eye movement, and then from right to left, N1, N2-N3, and N4 stages. Rapid eye movement overlaps with all stages. 

Figure \ref{fig:wesad_tsne} illustrates visualisations of raw signals from the WESAD dataset and their corresponding embedding vectors. The plots represent four different emotions including \textit{neutral}, \textit{stress}, \textit{amusement}, and \textit{meditation}. As Figure \ref{fig:wesad_tsne_raw} shows, the raw signals have formed five disjoint clusters regardless of their emotion labels. Similarly, extracted embeddings of each emotion have formed multiple condensed clusters (Figures \ref{fig:wesad_tsne_ssl} and \ref{fig:wesad_tsne_fine}). The most dense clusters are related to \textit{stress} and \textit{meditation} states, while \textit{neutral} and \textit{amusement} states  have many in common and are distributed across the latent space. These findings confirm the observation in \cite{deldari2020espresso} which propose an unsupervised segmentation method to extract human emotion. While emotions like \textit{stress} and \textit{meditation} are easily detectable, the others are harder to distinguish. This may root in the differences in personality traits and characteristics of individuals. For example, each person can have a different taste of amusing topics. This visualisation shows the importance of training subject-based models along with training the core model to capture user-specific features of biosignals as discussed in \cite{cheng2020subject}.

\subsection{Discussion and Future Works}
\label{sec:discussion}
We presented the effectiveness of our proposed objective function in learning compact and informative representations for several sensor modalities. The result from \textit{CMC} and \METHOD\ show that contrasting all modalities can be more beneficial and \METHOD\ provides superior improvement compared to the other multimodal objective function, \textit{CMC}. Although \textit{CMC} considers full combinatorial pair of modalities for contrastive loss function (which makes it more complex in terms of the number of modalities), \METHOD\ can still outperform it with linear time and space complexity in terms of the number of available modalities. In contrast to the limited number of modalities in the Computer vision area (usually limited to audio, vision, and text), the number of input sources in ubiquitous sensor applications is much more which introduces specific challenges and higher complexity. 

The main challenges in application with a various number of sensors include:
\begin{itemize}
    \item Data comes from different sensors with different noise and sampling rates. (\textbf{ch1})
    \item Not all sensors (data sources) contribute to all events and due to the dynamic nature of human-related tasks the contribution also changes during time. (\textbf{ch2})
    \item Data from some of the sensors may be missing at some points. (\textbf{ch3})
\end{itemize}
To address the first challenge (\textbf{ch1}) and following the recent work in multi-device contrastive learning \cite{jain2022collossl} we assumed, first, the small amount of noise and differences in sensors do not degrade the performance. Second, all input sensors share the same sampling rate or the data can be synchronised and re-sampled to the same rate.

despite the fact that data may come from many different sources, not all of them necessarily take part in every class of data. For example, considering the human activity recognition problem, an ankle-worn accelerometer sensor provides less amount of information toward ``\textit{eating}'' activity compared to the sensors positioned on either arms, head or ears. Considering irrelevant data may degrade the performance of the model. Authors in \cite{jain2022collossl} proposed to find the most relevant sensors dynamically by calculating the Maximum Mean Discrepancy (MMD) between each pair of sensors (they only considered accelerometer and gyroscope sensors). However, comparing every pair of available sensors frequently across the whole period can be a time-consuming task and grows in a combinatorial order regarding the number of sensors. On the other hand, sensors participating in a specific activity do not necessarily share similar distribution characteristics due to the different \textit{type} of the sensors and different \textit{positions} on the human body (e.g. SLEEPEDF and WESAD dataset). In this work, in each dataset, all available sensors contribute to almost all classes. Hence we aim to expand this work to include effective and dynamic sensor selection to consider changes in the rate of sensors' participation and relevancy (\textbf{ch2}). 

Our proposed method already covers the last but not least challenge (\textbf{ch3}) in the ubiquitous applications. Since \METHOD\ employs modality-specific encoders, the downstream task is not limited to the availability of all sensors. Proper encoders can be selected based on the type of sensors for training and inference of the downstream task.

Compared to the significant works on self-supervised learning in the computer vision and natural language processing fields, time-series data has been studied far less in SSL and CL-related literature. Therefore, although we do not consider any limiting assumption on the type and modality of input data, we focused on contrasting time-series data for pervasive and ubiquitous computing applications.
This research reveals some differences in applying existing self-supervised objective functions (which were mostly introduced for computer vision tasks) to time series. For example, we showed the effect of batch size to learn distinguishing features across different data sources. As we discussed in the previous section, unlike image-based applications, smaller batch sizes seemed to learn more effective representations, given they contained fewer false-negative samples in each batch. 



\section{Conclusion}
\label{sec:conclusion}
In this work, we focused on a cross-view contrastive learning approach for self-supervised representation learning from multi-sourced time-series data. Although our proposed method is modality-agnostic, we specifically focused on wearable and bio-sensors data. In contrast to existing self-supervised representation learning methods that focus only on one type of sensor (modality), We hypothesized that not only important contextual information is shared across all modalities, but each modality provides a complementary view of the whole system and can enhance the quality of the representation vector. In this regard, we proposed \METHOD, Cross mOdality COntrastive leArning, as a novel and simple contrastive learning-based objective function applicable for self-supervised representation learning from multimodal data. \METHOD\ is more computationally efficient compared to both supervised and state-of-the-art self-supervised multimodal contrastive learning models for representation learning from data with multiple modalities/views. We examined our proposed method to learn the representations for different sensing modalities using five well-known publicly available datasets.

We performed extensive experiments to evaluate the effectiveness of \METHOD\ against supervised approaches and five well-known SSL contrastive and non-contrastive objective functions. We showed the superiority of \METHOD\ in learning useful representations for the downstream classification task. In addition, we investigated our approach in terms of class label efficiency and showed that \METHOD\ can outperform supervised baselines by training our model with far smaller amounts of labelled data. We discussed the limitations of current work and suggested future avenues to improve SSL models for ubiquitous applications.

We believe \METHOD\ opens up  avenues for exploring self-supervised techniques for multi source time-series data.
The next step is investigating the applicability of \METHOD\ for vastly different data modalities such as video, audio and text and making contextual aware \METHOD.


\begin{acks}
Authors would like to acknowledge the support from CSIRO Data61 Scholarship program (Grant number 500588), RMIT Research International Tuition Fee Scholarship and Australian Research Council (ARC) Discovery Project DP190101485.
\end{acks}

\bibliographystyle{ACM-Reference-Format}
\bibliography{main}

\appendix

\end{document}